\documentclass[twoside,11pt]{article}

\usepackage[preprint]{jmlr2e}
\usepackage{graphicx}
\usepackage{amsfonts}
\usepackage{amsmath, amssymb}
\usepackage{bm}
\usepackage{xcolor}
\usepackage{tikz}
\usepackage{subcaption}
\usetikzlibrary{positioning}
\usepackage[strict]{changepage}
\usepackage{url}
\usepackage{tabularx}

\newcommand{\ranvec}[1]{\bm{#1}}
\newcommand{\ranmat}[1]{\bm{\overline{#1}}}
\newcommand\numberthis{\addtocounter{equation}{1}\tag{\theequation}}
\newtheorem{ass}[theorem]{Assumption}
\newtheorem{rmk}[theorem]{Remark}

\DeclareMathOperator{\erf}{erf}
\DeclareMathOperator{\sgn}{sgn}

\ShortHeadings{Richer priors for infinitely wide MLPs}{}
\firstpageno{1}

\begin{document}

\title{Richer priors for infinitely wide multi-layer perceptrons}

\author{\name Russell Tsuchida \email s.tsuchida@uq.edu.au \\ \addr School of Information Technology and Electrical Engineering  \\ The University of Queensland
\\ Brisbane, QLD 4072, Australia \AND
\name Fred Roosta \email fred.roosta@uq.edu.au \\ \addr  School of Mathematics and Physics \\ The University of Queensland \\ Brisbane, QLD 4072, Australia \AND
\name Marcus Gallagher \email marcusg@uq.edu.au \\ \addr School of Information Technology and Electrical Engineering \\ The University of Queensland \\ Brisbane, QLD 4072, Australia}

\editor{}

\maketitle

\begin{abstract}
It is well-known that the distribution over functions induced through a zero-mean iid prior distribution over the parameters of a multi-layer perceptron (MLP) converges to a Gaussian process (GP), under mild conditions. We extend this result firstly to independent priors with general zero or non-zero means, and secondly to a family of partially exchangeable priors which generalise iid priors. We discuss how the second prior arises naturally when considering an equivalence class of functions in an MLP and through training processes such as stochastic gradient descent. 

The model resulting from partially exchangeable priors is a GP, with an additional level of inference in the sense that the prior and posterior predictive distributions require marginalisation over hyperparameters. We derive the kernels of the limiting GP in deep MLPs, and show empirically that these kernels avoid certain pathologies present in previously studied priors. We empirically evaluate our claims of convergence by measuring the maximum mean discrepancy between finite width models and limiting models. We compare the performance of our new limiting model to some previously discussed models on synthetic regression problems. We observe increasing ill-conditioning of the marginal likelihood and hyper-posterior as the depth of the model increases, drawing parallels with finite width networks which require notoriously involved optimisation tricks\footnote{Code available at \url{https://github.com/RussellTsuchida/RicherPriorsForMLPs.git}}.
\end{abstract}

\section{Introduction}
\subsection{Notation}
We begin by defining our main notation for the paper. Upper-case letters denote random variables. Random vectors and random matrices are boldface and boldface-with-overlines respectively. The first subscript extracts the row (element) of a matrix (vector). The second subscript extracts the column of a matrix. Parenthesised superscripts indicate the layer to which an object belongs. The number of neurons in layer $l$ is denoted $n^{(l)}$, with $n^{(0)}$ referring to the input dimensionality. The activation function is denoted $\psi: \mathbb{R} \to \mathbb{R}$, with $\bm{\psi}: \mathbb{R}^d \to \mathbb{R}^d$ representing $\psi$ applied element-wise. Random activation signals in layer $l$ for inputs $\ranvec{x}$ are denoted $\ranvec{X}^{(l)}(\ranvec{x})$. Where unambiguous, we use the shorthand $\ranvec{X}^{(l)}$ and $\ranvec{X'}^{(l)}$ to denote  $\ranvec{X}^{(l)}(\ranvec{x})$ and $\ranvec{X}^{(l)}(\ranvec{x'})$. Subscripts on expectations $\mathbb{E}_Q$ indicate that the expectation is to be taken over the distribution of the random element $Q$. The absence of subscripts on expectations indicates that the expectation is over the joint distribution of all random variables. 

We will use several special functions throughout. $\Theta: \mathbb{R} \to \{0, 1/2, 1\}$, $\sgn: \mathbb{R} \to \{-1, 1\}$ and $\erf:\mathbb{R} \to [-1, 1]$ denote the Heaviside step, sign and error functions respectively. $\phi$ and $\Phi$ denote the PDFs and CDFs of the standard Gaussian random variable, and $\phi_2$ and $\Phi_2$ the bivariate counterparts.

\def\layersep{2.2cm}
\definecolor{mygreen}{rgb}{0.176, 0.718, 0.043}
\begin{figure}
\centering
\begin{tikzpicture}[shorten >=1pt,->,draw=black!50, node distance=\layersep]
    \tikzstyle{every pin edge}=[<-,shorten <=1pt]
    \tikzstyle{neuron}=[circle,fill=black!25,minimum size=25pt,inner sep=0pt]
    \tikzstyle{input neuron}=[neuron, fill=white, draw=black];
    \tikzstyle{output neuron}=[neuron, fill=white, draw=black];
    \tikzstyle{hidden neuron}=[neuron, fill=white, draw=black];
    \tikzstyle{annot} = [text width=4em, text centered]

    % Draw the input layer nodes
    \foreach \name / \y in {1,...,4}
    % This is the same as writing \foreach \name / \y in {1/1,2/2,3/3,4/4}
        %\node[input neuron, pin=left:Input \#\y] (I-\name) at (0,-\y) {};
        \node[input neuron] (I-\name) at (0,-\y) {$x_\name$};
        
    % Draw the hidden layer nodes
    \foreach \name / \y in {1,...,5}
        \path[yshift=0.5cm]
            node[hidden neuron] (H1-\name) at (\layersep,-\y cm) {$X^{(1)}_\name$};
            
    \foreach \name / \y in {1,...,5}
        \path[yshift=0.5cm]
            node[hidden neuron] (H2-\name) at (2*\layersep,-\y cm) {$X^{(2)}_\name$};

    % Draw the output layer node
    \node[output neuron, right of=H2-3] (O-1) {$f(\ranvec{x})$};

    % Connect every node in the input layer with every node in the
    % hidden layer.
    \foreach \source in {1,...,4}
        \foreach \dest in {1,...,5}{
            \path (I-\source) edge (H1-\dest);}
            
    \foreach \source in {1,...,4}{
    	\path[every node/.style={sloped,anchor=north,auto=false}] (I-\source) edge[blue] (H1-5);}
    \node[annot, text=blue, below right=0.1cm and 0.1cm of I-4] {$\ranvec{W}^{(1)}_5$};
            
    \foreach \dest in {1,...,5}{
        \foreach \destt in {1,...,5}
            \path (H1-\dest) edge (H2-\destt);}
            
   	\path[every node/.style={anchor=north,auto=false}] (H1-5) edge[red] node {$W^{(2)}_{45}$} (H2-4);

    % Connect every node in the hidden layer with the output layer
    \foreach \source in {1,...,5}
        \foreach \dest in {1,...,1}{
            \path (H2-\source) edge (O-\dest);}
    \path[every node/.style={sloped,anchor=south,auto=false}] (H2-3) edge[mygreen] node {$W^{(3)}_{3}$} (O-1);

    % Annotate the layers
    %\node[annot,above of=I-1, node distance=1.5cm] (hl) {$n^{(0)}$};
    %\node[annot,right of=hl] (hmid) {$n^{(1)}$};
    %\node[annot,right of=hmid] (hlast) {$n^{(2)}$};
    %\node[annot,right of=hlast] {$n^{(3)}$};

    % Weights
    \node[annot, above right=0.3cm and 0.2cm of I-1] (W1) {$\ranmat{W}^{(1)}$};
    \node[annot,right of=W1] (W2) {$\ranmat{W}^{(2)}$};
    \node[annot,right of=W2] (W3) {$\ranvec{W}^{(3)}$};
\end{tikzpicture}
\caption{Example MLP showing notation.}
\end{figure}
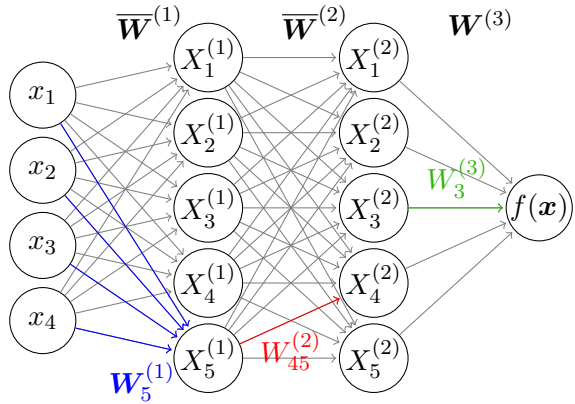

\subsection{Background}
\subsubsection{MLPs with iid weights as GP priors}
\label{sec:iid_gp}
A classical result connecting GPs with MLPs is given by Neal~\citeyearpar{neal1995bayesian}, which we briefly review here. Consider a $1$-hidden layer MLP with one output\footnote{This description may be extended to $n^{(2)}$ outputs, either by assuming independence between outputs, or through some more involved method~\citep[p. 190]{RW_GP}.}, random input-to-hidden weights\footnote{To simplify the exposition of existing work, we ignore biases. Biases are handled in~\S\ref{sec:single_layer_kernel}.} $\ranmat{W}^{(1)}$ containing iid entries, and random hidden-to-output weights $\ranvec{W}^{(2)}$ with mean $0$ and variance $\frac{1}{n^{(1)}}$. Furthermore, take $\ranmat{W}^{(1)}$ to be independent of $\ranvec{W}^{(2)}$. The MLP defines the random function $f(\ranvec{x}) = \ranvec{W}^{(2)} \cdot \bm{\psi}(\ranmat{W}^{(1)} \ranvec{x})$. The distribution over the weights induces a distribution over functions $f$. The mean $\mathbb{E}\big[f(\ranvec{x})\big]$ is zero for all $\ranvec{x}$, since elements of $\ranvec{W}^{(2)}$ have zero mean and are independent of $\ranmat{W}^{(1)}$. The covariance function is given by
\begin{align*}
    k(\ranvec{x}, \ranvec{x'}) :&= \mathbb{E}[f(\ranvec{x})f(\ranvec{x'})] -  \mathbb{E}[f(\ranvec{x})]\mathbb{E}[f(\ranvec{x'})]\\
    &=\mathbb{E} \Big[ \psi(\ranvec{W}^{(1)}_1 \cdot \ranvec{x}) \psi(\ranvec{W}^{(1)}_1 \cdot \ranvec{x'}) \Big] \numberthis \label{eq:kernel_def}. 
\end{align*}
As the output is a sum of independent random variables, the joint distribution of $f$ over any finite collection of $\{\ranvec{x}_i\}_{i=1}^N$ approaches a multivariate Gaussian as $n^{(1)} \to \infty$ due to the central limit theorem (CLT), assuming suitably well-behaved weight distributions and $\psi$. While the application of the CLT is not formally handled with respect to conditions on the weight distributions and activation functions in the original work~\citep{neal1995bayesian}, it has been applied more rigourously in recent work~\citep{matthews2018gaussian, yang2019scaling, NIPS2019_9186}.

Following the usual presentation of GPs~\citep{RW_GP}, for any finite collection of $N$ points $\ranmat{X}^{(0)} \in \mathbb{R}^{N \times n^{(0)}},$ 
\begin{equation}
    \lim\limits_{n^{(1)} \to \infty}\ranvec{f}\big(\ranmat{X}^{(0)}\big) \sim \mathcal{N}\Big(\ranvec{0}, K\big(\ranmat{X}^{(0)}, \ranmat{X}^{(0)}\big)\Big) \label{eq:gp_prior},
\end{equation}
where $K\big(\ranmat{X}^{(0)}, \ranmat{X}^{(0)}\big)$ denotes the matrix obtained by applying $k$ to every tuple in $\ranmat{X}^{(0)}$. We are often interested in making predictions over new data $\ranmat{X}_*$ given observations of training targets modelled by $\ranvec{y} = \lim\limits_{n^{(1)} \to \infty} \ranvec{f}\big( \ranmat{X}^{(0)} \big) + \ranvec{Z}$, where $\ranvec{Z}$ is isotropic Gaussian noise\footnote{Non-trivial noise structures are also possible~\citep[p. 191]{RW_GP}.} independent of $\ranvec{f}$ with variance $s^2 \geq 0$. To do this, we use the posterior predictive distribution, obtained through conditioning. Letting $\ranvec{f}_* = \lim\limits_{n^{(1)} \to \infty} \ranvec{f}\big(\ranmat{X}_*\big)$, the posterior predictive distribution is
\begin{align*}
    &\ranvec{f}_* | \ranmat{X}_*, \ranmat{X}^{(0)}, \ranvec{y} \sim \mathcal{N}\big( \ranvec{\overline{f}}_*, \, \text{cov}(\ranvec{\overline{f}}_*) \big), \quad \text{where} \numberthis \label{eq:gp_pred} \\
    \ranvec{\overline{f}}_* &= K(\ranmat{X}_*, \ranmat{X}^{(0)}) [K(\ranmat{X}^{(0)},\ranmat{X}^{(0)}) + s^2 I]^{-1} \ranvec{y}, \quad \text{and} \\
    \text{cov}(\ranvec{\overline{f}}_*) & = K(\ranmat{X}_*, \ranmat{X}_*) -  K(\ranmat{X}_*, \ranmat{X}^{(0)}) \big[ K(\ranmat{X}^{(0)},\ranmat{X}^{(0)})+s^2I \big]^{-1}K(\ranmat{X}^{(0)},\ranmat{X}_*). 
\end{align*}
The prior~\eqref{eq:gp_prior} and posterior predictive~\eqref{eq:gp_pred} are defined completely by the kernel~\eqref{eq:kernel_def}, which in turn depends on the choice of activation function $\psi$ and distribution over hidden weights. Closed form expressions for the kernel for some different choices of activation functions and weight distributions are known as described in Table~\ref{tab:analytic_kernels}.

\setlength\tabcolsep{3pt}
\begin{table*}[!t]
\centering
\begin{tabularx}{\linewidth}{l|l|l|l}
    $\bm{\psi(z)}$ & \textbf{Name}  & \textbf{Weight distribution} & \textbf{Source} \\
                  \hline
	$\phi(z)$ & Error function & $0$-mean Gaussian &  \cite{williams1997computing}  \\ \hline
	-\footnotemark & Gaussian & $0$-mean Gaussian &  \cite{williams1997computing}  \\ \hline
		$\sgn(z)$ &Sign function & $0$-mean uniform &  \cite{le2007continuous} \\ \hline
                  $\Theta(z) z^n$ & Step, $n=0$ & $0$-mean Gaussian  &\cite{sheppard1899iii, o2014analysis} \\
          &       ReLU, $n=1$  &&  \cite{bibi2018analytic} \\
          & General $n$ & & \cite{NIPS2009_3628} \\ \hline
        $\Theta(z) z + a\Theta(-z)z$ &LReLU &  $0$-mean Gaussian &  \cite{tsuchida2018invariance} \\ \hline
        $\sin(az)$ or $\cos(az)$ & Trigonometric & Gaussian  & \cite{pearce2019expressive} \\
\end{tabularx}
\caption{Some known single-layer kernels of networks with given weight distributions and activation functions. The derivations of some of these kernels have motivations unrelated to our current focus.}
\label{tab:analytic_kernels}
\end{table*}
\footnotetext{This activation function is applied in Radial Basis Function networks, differing from our current setup in that the layer does not perform a matrix multiplication followed by a non-linearity. We include it here for completeness, although the setup is not directly comparable.}

Recent extensions of the above discussion to deep models are discussed in concurrent work of~\cite{lee2017deep} and~\cite{matthews2018gaussian}. Roughly, when the priors are iid with zero mean, the model still converges to a GP with a kernel in the final layer that may be defined recursively as a function of the kernel in the previous layer. This kernel may be evaluated exactly or numerically for specific choices of the activation function. Other network architectures have also been previously studied, namely convolutional architectures~\citep{garriga2018deep, novak2018bayesian} and general compositions of recurrent, graph convolution, pooling, skip connection, attention and normalisation layers~\citep{NIPS2019_9186, yang2019scaling}. 

The focus of our work is not to study the effect of the architecture on the kernel, but rather to study the effect of the parameter distributions on the kernel and more generally, the model.

\subsubsection{A fixed point in deep models with $0$-mean iid weight priors and LReLU activations}
% CTRL + F low_res_figures for regular vectorised pdf figures when submitting non-arXiv version
\begin{figure*}[t!]
\centering
\begin{minipage}[t]{.24\linewidth}
\includegraphics[width=\linewidth]{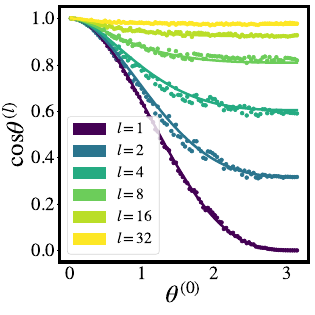}
\subcaption{}
\end{minipage}\hfill
\begin{minipage}[t]{.24\linewidth}
\includegraphics[width=\linewidth]{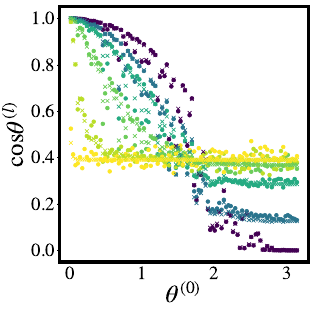}
\subcaption{}
\end{minipage}
\begin{minipage}[t]{.24\linewidth}
\includegraphics[width=\linewidth]{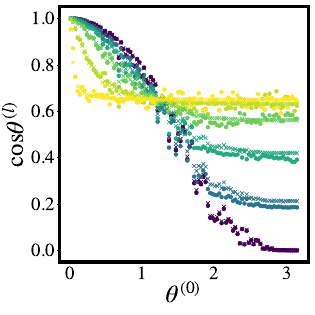}
\subcaption{}
\end{minipage}
\begin{minipage}[t]{.24\linewidth}
\includegraphics[width=\linewidth]{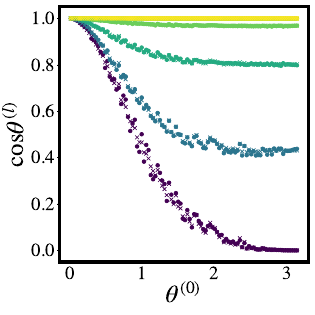}
\subcaption{}
\end{minipage}
    \caption{Theoretical and empirical normalised kernels of $l$ layer MLPs with ReLU activations, $2$ inputs, $3000$ hidden units in each layer and independent Gaussian weights according to~\eqref{eq:iid_prior}. Inputs are generated through mutual random rotations of $(1,0)^\top$ and $(\cos\theta^{(0)},\sin\theta^{(0)})^\top$. Theoretical predictions are shown by the curve in (a), and $\times$ in (b), (c), (d). Empirical samples are shown by $\bullet$. Parameterisation according to~\eqref{eq:iid_prior}: (a) $(\mu^{(l)}_{ij}, \sigma^{(l)}_{ij}) = (0, \sqrt{2})$, (b) $(\mu^{(l)}_{ij}, \sigma^{(l)}_{ij}) = (-2, \sqrt{2})$, (c) $(\mu^{(l)}_{ij}, \sigma^{(l)}_{ij}) =(-1, \sqrt{2})$, (d) $(\mu^{(l)}_{ij}, \sigma^{(l)}_{ij}) = (1, \sqrt{2})$. \textbf{Note:} In all but (a), the theoretical kernels are not functions of only $\theta^{(0)}$, but of $\ranvec{x}$ and $\ranvec{x}'$. Plotting samples in this manner merely provides a useful visualisation, and in particular shows good agreement between asymptotic theory ($\times$) and finite width samples ($\bullet$).}
    \label{fig:iid_kernel}
\end{figure*}
\label{sec:pathology}
Taking an MLP with Leakly Rectified Linear Unit (LReLU) activations $\phi(z)=\max(az, z), a \in (-1, 1)$ and an independent Gaussian prior having mean $0$ and variance $(\sigma^{(l)})^2/n^{(l-1)}$ for each layer, the covariance in layer $L$ can be found iteratively~\citep{tsuchida2018invariance}:
\begin{align*}
    k \big( \ranvec{x}, \ranvec{x'} \big) &= c^2 \Vert \ranvec{x} \Vert \Vert \ranvec{x'} \Vert \cos\theta^{(L)}, \\
    \cos\theta^{(l)} &= \frac{(1-a)^2}{\pi(1+a^2)} \big( \sin\theta^{(l-1)} + (\pi-\theta^{(l-1)})\cos\theta^{(l-1)} \big) + \frac{2a}{1+a^2}\cos\theta^{(l-1)}, \quad 1 \leq l \leq L,\\
    \cos\theta^{(0)} &= \frac{\ranvec{x}\cdot \ranvec{x'}}{\Vert \ranvec{x} \Vert \Vert\ranvec{x'} \Vert},\numberthis \label{eq:arc-cosine}
\end{align*}
where\footnote{Note the reparameterisation of hyperparameters; the product of variances in each layer collapses into a single hyperparameter.} $c^2 = \frac{(1+a^2)^L}{n^{(0)}2^L} \prod_{l=1}^L (\sigma^{(l)})^2$. This kernel generalises the arc-cosine kernel of degree $1$~\citep{NIPS2009_3628} where $a=0$. In both infinitely wide GPs~\citep{NIPS2016_6427}, and finitely wide MLPs~\citep{he2015delving}, when $a=0$ it is typical to set $(\sigma^{(l)})^2 = 2$ to ensure that $c^2$ is constant in $L$. More generally when $a\neq 0$, setting $(\sigma^{(l)})^2 = 2/(1+a^2)$ keeps $c^2$ constant in $L$~\citep{tsuchida2018invariance}. Our analysis relates to the general LReLU $a \in (-1, 1)$, but for simplicity we restrict our illustrations and experiments to $a=0$ throughout this paper.

The normalised kernel resulting from~\eqref{eq:arc-cosine}, shown in Figure~\ref{fig:iid_kernel}(a) and depending only on $\theta^{(0)}$, reveals a potentially undesirable phenomenon arising from the use of $0$-mean iid priors on the weights on an MLP. As the signal propagates through the layers, the covariance function approaches $c^2 \Vert \mathbf{x} \Vert \Vert \mathbf{x}' \Vert$, gradually losing dependence on $\theta^{(0)}$. This means that function draws will look like constant multiples of $\Vert \mathbf{x} \Vert$ as $L$ becomes large. 

\begin{figure*}[!t]
\centering
\begin{minipage}[t]{.195\linewidth}
\includegraphics[width=\linewidth]{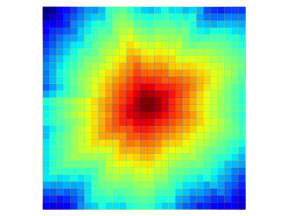}
\subcaption{$(0, \sqrt{2})$}
\end{minipage}\hfill
\begin{minipage}[t]{.195\linewidth}
\includegraphics[width=\linewidth]{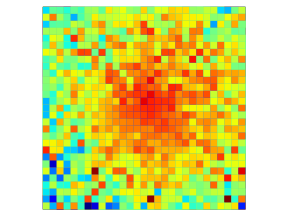}
\subcaption{$(-0.3, 1.52)$}
\end{minipage}
\begin{minipage}[t]{.195\linewidth}
\includegraphics[width=\linewidth]{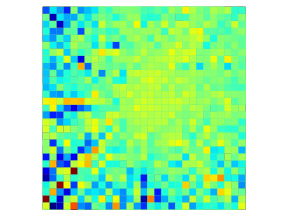}
\subcaption{$(-0.65, 1.63)$}
\end{minipage}\hfill
\begin{minipage}[t]{.195\linewidth}
\includegraphics[width=\linewidth]{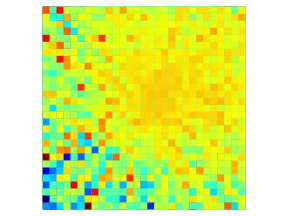}
\subcaption{$(-1.05, 1.75)$}
\end{minipage}
\begin{minipage}[t]{.195\linewidth}
\includegraphics[width=\linewidth]{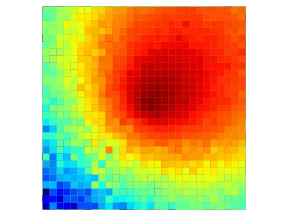}
\subcaption{$(-1.51, 1.89)$}
\end{minipage}
\centering
\begin{minipage}[t]{.195\linewidth}
\includegraphics[width=\linewidth]{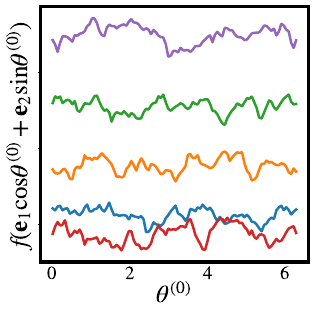}
\subcaption{}
\end{minipage}\hfill
\begin{minipage}[t]{.195\linewidth}
\includegraphics[width=\linewidth]{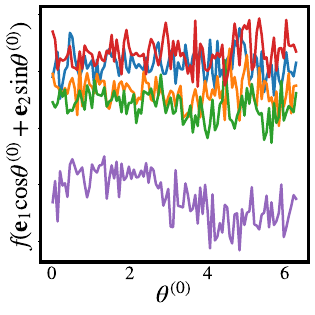}
\subcaption{}
\end{minipage}
\begin{minipage}[t]{.195\linewidth}
\includegraphics[width=\linewidth]{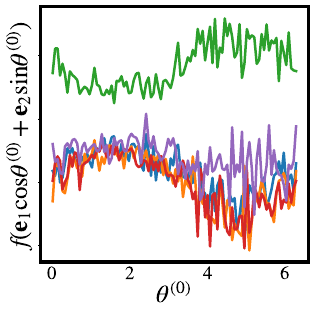}
\subcaption{}
\end{minipage}\hfill
\begin{minipage}[t]{.195\linewidth}
\includegraphics[width=\linewidth]{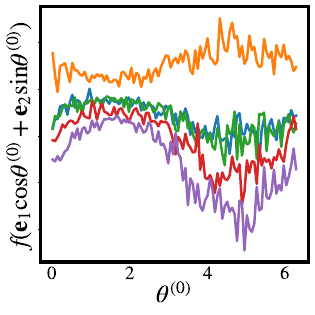}
\subcaption{}
\end{minipage}
\begin{minipage}[t]{.195\linewidth}
\includegraphics[width=\linewidth]{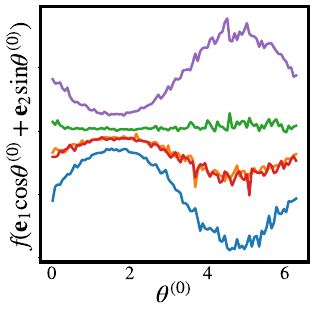}
\subcaption{}
\end{minipage}
\caption{ \textbf{(a - e)} $1$ sample draw from the limiting GP of an MLP with random weights and $2$ inputs, $64$ hidden layers and one output. The subcaption shows the values of $(\mu^{(l)}_{ij}, \sigma^{(l)}_{ij})$ according to~\eqref{eq:iid_prior}. \textbf{(f - j)} $5$ sample draws from the limiting GP of the same MLP but with $10$ inputs. In order to visualise the function, we choose two random orthogonal directions $\mathbf{e}_1$ and $\mathbf{e}_2$, and then traverse the unit hypersphere on the circle defined by these two directions. The directions are generated through a QR decomposition of a matrix with iid standard Gaussian elements.}
\label{fig:iid_draws_2d}
\end{figure*}

In Figure~\ref{fig:iid_draws_2d}(a), we show a sample function draw from the limiting GP of an MLP over a $2$ dimensional input, which looks like a constant multiple of $\Vert \mathbf{x} \Vert$. In Figure~\ref{fig:iid_draws_2d}(f), we show $5$ sample function draws from the limiting GP of an MLP over a $10$ dimensional input. Since the covariance function becomes degenerate (corresponding to a \emph{constant} covariance matrix) as the depth becomes large, the function draws are almost constant on the unit circle. 

Another way to understand this phenomenon is as follows. We may bound the probability that two points of a single sample of a GP prior differ by more than an amount relative to the scale of the problem given by the hyperparameter $c$ in~\eqref{eq:arc-cosine}. When the inputs have unit norms, the variance of $f(\ranvec{x}_1) - f(\ranvec{x}_2)$ is $2 \sigma^2(1-\cos\theta^{(L)})$. Chebyshev's inequality then gives
\begin{align*}
\mathbb{P}\Bigg( \big| f(\ranvec{x}_1) - f(\ranvec{x}_2) \big| \geq \sqrt{2} \sigma k \sqrt{1-\cos\theta^{(L)}} \Bigg) \leq \frac{1}{k^2}
\end{align*} 
for any real $k > 0$. As observed in Figure~\ref{fig:iid_kernel} (a), $\cos\theta^{(L)}$ approaches the fixed point of $1$ as $L$ increases. This has been proven by different authors in different contexts, notations and levels of generality~\citep{NIPS2016_6322, tsuchida2018invariance}.

The observation that the covariance function associated with $0$-mean iid weight priors exhibits this behaviour motivates research into an increased understanding of the effects of prior parameterisations on infinitely wide MLPs.

\subsubsection{Convergence occurs under restrictive assumptions}
The mathematical correspondance between MLPs and GPs may seem at odds with recently observed empirical performance. While GPs are mathematically elegant and principled, MLPs outperform them on many tasks. Furthermore, depth is commonly touted theoretically~\citep{NIPS2016_6322, mhaskar2017and} and empirically~\citep{NIPS2012_4824, simonyan2014very, szegedy2015going, he2016deep} as a means of increasing the expressivity and performance of neural networks. If deep MLPs perform so well, why do deep infinitely wide MLPs suffer from increasingly ``simple" kernels?

These observations can be explained through the hugely restrictive prior that is used to guarantee convergence. There are two very obvious mathematical conveniences used in this prior: the scaling of the weights and the iid assumption. Relaxing these conveniences, a more interesting question is perhaps 

\begin{center}
\textit{``When do MLPs \emph{not} converge to GPs, but still converge to something conceptually if not computationally tractable?"}
\end{center}

Related questions have been posed by multiple authors~\citep{neal1995bayesian, mackay2003information, matthews2018gaussian, chizat2019lazy}. We make two theoretical contributions in this direction.

\subsection{Contributions}
Firstly, we show that networks with non-zero mean iid weights can also converge to GPs. We analytically derive the kernels of deep models with non-zero mean iid weights. Theoretical and empirical normalised kernels of these networks are shown in Figure~\ref{fig:iid_kernel}(b), (c), and (d). As observed in Figure~\ref{fig:iid_kernel}(b), certain non-zero means can alleviate the pathological behaviour in Figure~\ref{fig:iid_kernel}(a). Sample function draws from the limiting GP of an MLP with non-zero means are shown in Figure~\ref{fig:iid_draws_2d}. This first contribution is a pre-requisite for our more general second contribution.

Secondly, we relax the iid prior to a partially exchangeable prior which allows for dependence inside layers but not between layers. In this case, the network converges to a GP with an additional layer of inference; instead of taking a single set of kernel hyperparameters, the limiting model requires integration over a hyper-prior.

We provide empirical verification of our theoretical results and apply our limiting models to some benchmark problems. We compare the performance of models with zero-mean parameter priors, non-zero mean parameter priors, and marginalised hyper-priors.

\section{Non-zero mean priors}
\label{sec:nonzero_mean}
We firstly evaluate the kernel~\eqref{eq:kernel_def} of a one hidden layer network with independent Gaussian parameters. Then we show convergence of deep networks with independent parameters to a GP, with a kernel defined iteratively.

\subsection{Single-layer kernels}
\label{sec:single_layer_kernel}
Suppose each row of the random weight matrix has the same distribution ($\ranvec{W}^{(1)}_{j_1} \stackrel{d}{=} \ranvec{W}^{(1)}_{j_2}$ $\forall j_1, j_2$) and each bias has the same distribution. Without ambiguity, we temporarily supress the layer superscripts and node subscripts for readability (for example, $\ranvec{W}^{(1)}_j$ is the same as $\ranvec{W}$). 

Let $\ranvec{Z}_W \sim \mathcal{N}(\ranvec{0}, I)$, $Z_b \sim \mathcal{N}(0,1)$ and $\ranvec{Z} = (\ranvec{Z}_w^\top, Z_b)^\top$. Let $\ranvec{W} = \Sigma^{(1/2)} \ranvec{Z} + \bm{\mu}$, where $\Sigma^{(1/2)}$ is a diagonal matrix with non-negative entries $\sigma_i$. Let $\ranvec{x}_1$ and $\ranvec{x}_2$ be network inputs, and let $\hat{\ranvec{x}}_i =  (\ranvec{x}_i^\top, 1)^\top$. This augmentation is performed to handle biases. Let $X = \psi(\ranvec{W} \cdot \hat{\ranvec{x}}_1)$ and $X' = \psi(\ranvec{W} \cdot \hat{\ranvec{x}}_2)$. Before tackling LReLU activations, we first evaluate the kernel for linear and absolute value activations.

\subsubsection{Linear activations}
Let $\theta=\cos^{-1}\frac{\hat{\ranvec{x}}_1^\top \Sigma \hat{\ranvec{x}}_2}{\Vert \Sigma^{(1/2)}\hat{ \ranvec{x}}_1 \Vert \Vert \Sigma^{(1/2)} \hat{\ranvec{x}}_2 \Vert}$ be the angle between $\hat{\mathbf{x}}_1$ and $\hat{\mathbf{x}}_2$ in space stretched by $\Sigma^{(1/2)}$. Taking $\psi$ to be the identity, the kernel $\mathbb{E}\big[ X X' \big]$ is given by
\begin{align*}
 &\phantom{{}={}}\mathbb{E}\big[ (\ranvec{Z} \cdot \Sigma^{(1/2)} \hat{\mathbf{x}}_1) (\ranvec{Z} \cdot \Sigma^{(1/2)} \hat{\mathbf{x}}_2)\big] + (\bm{\mu} \cdot \hat{\mathbf{x}}_1) (\bm{\mu} \cdot \hat{\mathbf{x}}_2 ) \\
 &= \Vert \Sigma^{(1/2)}  \hat{\mathbf{x}}_1 \Vert \Vert \Sigma^{(1/2)}  \hat{\mathbf{x}}_2 \Vert \mathbb{E} \big[ Z_1(Z_1\cos\theta + Z_2\sin\theta) \big] + (\bm{\mu} \cdot \hat{\mathbf{x}}_1) (\bm{\mu} \cdot \hat{\mathbf{x}}_2 )\\
&= \Vert \Sigma^{(1/2)}  \hat{\mathbf{x}}_1 \Vert \Vert \Sigma^{(1/2)}  \hat{\mathbf{x}}_2 \Vert \cos\theta + (\bm{\mu} \cdot \hat{\mathbf{x}}_1) (\bm{\mu} \cdot \hat{\mathbf{x}}_2 ) \label{eq:lin_kernel} \numberthis.
\end{align*}
In order to go from the first line to the second line, we use rotational-invariance of the standard Gaussian distribution. There exists an orthogonal matrix $R$ such that $\Sigma^{(1/2)} \hat{\mathbf{x}}_1=\Vert \Sigma^{(1/2)}  \hat{\mathbf{x}}_1 \Vert R(1,0,...0)^\top $ and $\Sigma^{(1/2)} \hat{\mathbf{x}}_2=\Vert \Sigma^{(1/2)}  \hat{\mathbf{x}}_1 \Vert R(\cos\theta,\sin\theta,0,...0)^\top$. After using this representation, we replace $\ranvec{Z} \stackrel{d}{=} R \ranvec{Z}$ inside the expectation. Finally, the dot product is invariant to rotations by $R$. We will use this ``rotation trick" again in this section.

\subsubsection{Absolute value activations}
The joint distribution of $(|\ranvec{W} \cdot \hat{\ranvec{x}}_1|, |\ranvec{W} \cdot \hat{\ranvec{x}}_2|)$ is described by a folded Gaussian distribution, for which much analysis exists~\citep{kan2017moments}. In particular, the cross-moments can be stated in terms of Gaussian PDFs, CDFs and the error function.

Let $\Phi(\cdot)$ denote the CDF of a univariate Gaussian distribution having zero mean and unit variance. Let $\Phi_2(\cdot, \cdot; \rho)$ denote the CDF of a bivarate Gaussian distribution having zero mean components, unit variances, and covariances $\rho$. Let $\phi$ and $\phi_2$ denote analogous PDFs. Then~\citep{kan2017moments}
\begin{align*}
 \mathbb{E}|\ranvec{W} \cdot \hat{\ranvec{x}}_1||\ranvec{W} \cdot \hat{\ranvec{x}}_2| &=\Big( \tilde{\sigma}_1\tilde{\sigma}_2 \tilde{\mu}_1 \tilde{\mu}_2 + \tilde{\sigma}_1\tilde{\sigma}_2 \cos\theta \Big)\big( 4 \Phi_2(\tilde{\mu}_1, \tilde{\mu}_2; \cos\theta) - 2\Phi(\tilde{\mu}_1) - 2\Phi(\tilde{\mu}_2) + 1 \big) \\
&\phantom{{}={}}+ 2 \tilde{\mu}_1 \tilde{\sigma}_1 \tilde{\sigma}_2 \phi(\tilde{\mu}_2) \erf \Big( \frac{\tilde{\mu}_1 - \cos\theta \tilde{\mu}_2}{\sqrt{2}\sin\theta}\Big) \\
&\phantom{{}={}}+ 2\tilde{\mu}_2 \tilde{\sigma}_1 \tilde{\sigma}_2 \phi(\tilde{\mu}_1) \erf \Big( \frac{\tilde{\mu}_2 - \cos\theta \tilde{\mu}_1}{\sqrt{2}\sin\theta}\Big) \\
&\phantom{{}={}}+4 \tilde{\sigma}_1\tilde{\sigma}_2 \sin^2\theta \phi_2(\tilde{\mu}_1, \tilde{\mu}_2;\cos\theta), \numberthis \label{eq:abs_kernel}
\end{align*}
where $\tilde{\sigma}^2_i= \Vert \Sigma^{(1/2)} \mathbf{x}_i \Vert^2 $ and $\tilde{\mu}_i = \bm{\mu}\cdot \hat{\mathbf{x}}_i/\tilde{\sigma}_i$. It is now also convenient to state that
\begin{align*}
\mathbb{E}|\ranvec{W} \cdot \hat{\ranvec{x}}_i| &= \tilde{\sigma}_i\tilde{\mu}_i  \erf\Big( \frac{\tilde{\mu}_i}{\sqrt{2}} \Big) + 2 \tilde{\sigma}_i \phi(\tilde{\mu}_i), \numberthis \label{eq:folded_normal_mean}
\end{align*}
which will be used in the evaluation of integrals to follow.

\subsubsection{LReLU activations}
Let $\psi(z)=\max(az, z)=\frac{1}{2}\big((1+a)z+(1-a)|z|\big)$. The LReLU activation enjoys the absolute homogeneity property, that for all $s \in \mathbb{R}$,
\begin{equation}
\psi(|s|z) = |s| \psi(z), \label{eq:abs_homo}
\end{equation}
which will be referred to throughout this paper. We would like to evaluate the kernel
\begin{align*}
\frac{1}{4} \mathbb{E} \big[& (1+a)^2 (\ranvec{W} \cdot \hat{\ranvec{x}}_1)(\ranvec{W} \cdot \hat{\ranvec{x}}_2) + (1-a^2) (\ranvec{W} \cdot \hat{\ranvec{x}}_1)|\ranvec{W} \cdot \hat{\ranvec{x}}_2| + \\
& (1-a^2)(\ranvec{W} \cdot \hat{\ranvec{x}}_2) |\ranvec{W} \cdot \hat{\ranvec{x}}_1| \, \,+ (1-a)^2 |\ranvec{W} \cdot \hat{\ranvec{x}}_1||\ranvec{W} \cdot \hat{\ranvec{x}}_2|\big].
\end{align*}
The first and last terms are multiples of kernels~\eqref{eq:lin_kernel} and~\eqref{eq:abs_kernel} with linear and absolute value activations. The remaining terms are now considered. By symmetry, we need only evaluate one of them since the other just involves a permutation of the subscripts. Using the ``rotation trick", $\mathbb{E}\big[ (\ranvec{W} \cdot \hat{\ranvec{x}}_1)|\ranvec{W} \cdot \hat{\ranvec{x}}_2| \big]$ is
\begin{align*}
&\phantom{{}={}}   \mathbb{E} \Bigg[ \Big( Z_1 \cos\theta + Z_2 \sin\theta + \tilde{\mu}_1 \Big) \Big|Z_1 + \tilde{\mu}_2 \Big| \Bigg]\tilde{\sigma}_1 \tilde{\sigma}_2 \\
&= \Bigg( \cos\theta \mathbb{E} \big[|Z_1 + \tilde{\mu}_2| (Z_1 + \tilde{\mu}_2) \big] - \cos\theta \tilde{\mu}_2 \mathbb{E}\big| Z_1 + \tilde{\mu}_2 \big| + \tilde{\mu}_1 \mathbb{E}  \big| Z_1 + \tilde{\mu}_2 \big| \Bigg)\tilde{\sigma}_1 \tilde{\sigma}_2.
\end{align*}
The second two terms are readily evaluated using the mean of the folded Gaussian distribution~\eqref{eq:folded_normal_mean}. For the first term, let $\sgn$ denote the function assigning $-1$ to negative values, $0$ to $0$, and $1$ to positive values. Let $\Theta$ denote the Heaviside step function. $\Theta$ and $\sgn$ are related through $\sgn(z) = 2\big(\Theta(z)-0.5\big)$. The expectation in the first term may be written as
\begin{align*}
&\phantom{{}={}}\mathbb{E} \big[|Z_1 + \tilde{\mu}_2| (Z_1 + \tilde{\mu}_2) \big] \\
& = \mathbb{E} \big[\sgn(Z_1 + \tilde{\mu}_2) (Z_1 + \tilde{\mu}_2)^2 \big] \\
&= 2 \mathbb{E} \big[\Theta(Z_1 + \tilde{\mu}_2) (Z_1 + \tilde{\mu}_2)^2 \big] - \mathbb{E} \big[ (Z_1 + \tilde{\mu}_2)^2 \big].
\end{align*}
The second term is simply $1 + \tilde{\mu}_2^2$. Let $Q = Z_1 + \tilde{\mu}_2$. The expectation in the first term is given by
\begin{align*}
\mathbb{E} \big[\Theta(Q) Q ^2 \big] &= \mathbb{E} \big[ Q^2 \mid Q > 0 \big]\mathbb{P}(Q > 0) \\
&= \mathbb{E} \big[ Q^2 \mid Q > 0 \big] \Bigg( 1 - \Phi\Big( \frac{-\tilde{\mu}_2}{1} \Big) \Bigg) \\
&= (1 + \tilde{\mu}_2^2)\Bigg( 1 - \Phi\Big( -\tilde{\mu}_2 \Big) \Bigg) + \tilde{\mu}_2 \phi \Big( -\tilde{\mu}_2 \Big).
\end{align*}

We may now assemble the expressions to arrive at the kernel for the LReLU activation. We do not assemble the kernel here (due to it not being very insightful), but do provide a software implementation\footnote{\url{https://github.com/RussellTsuchida/RicherPriorsForMLPs.git}} in the Python library~\cite{gpy2014}. Evaluation of the kernel involves the bivariate Gaussian CDF, which is currently not efficiently vectorised in Python. Therefore, we use the FORTRAN routines of~\cite{genz2004numerical} packaged into a Python module using f2py~\citep{peterson2009f2py}.

\subsection{Deep networks converge to GPs}
Analytical justification for convergence to a GP follows the same arguments as in the work of previous authors~\citep{lee2017deep, matthews2018gaussian}, with two small modifications. Firstly, in our setup each column of the weight matrix is allowed to have a different mean and the diagonal covariance matrix has distinct entries. Secondly, in addition to requiring the variance of the weights to scale like $\frac{1}{n^{(l-1)}}$, we also require the means of the weights to scale like $\frac{1}{n^{(l-1)}}$. Formally,
\begin{align*}
W^{(l)}_{ji} &= \frac{1}{\sqrt{n^{(l-1)}}}\Big( \sigma_{i}^{(l)} Z^{(l)}_{ji} + \frac{\mu^{(l)}_{i}}{\sqrt{n^{(l-1)}}} \Big) \numberthis \label{eq:iid_prior},
\end{align*}
where each $Z^{(l)}_{ji}$ is a standard Gaussian, independent from one another. In order to simplify notation and conceptualisation, we will ignore biases for the remainder of the paper (i.e. we set all biases to $0$). In future work, biases may be handled in the same manner as in~\S\ref{sec:single_layer_kernel}, by appropriately augmenting the weight and input vectors.

We take the width of each layer to infinity, starting with the first layer and then continuing in order with subsequent layers. Under this scheme, it is reasonable to expect that the contributions of the deterministic mean components in the pre-activations converge by a law of large numbers (LLN), and the random components converge by a central limit theorem (CLT). Mirroring~\cite{lee2017deep}, we make the following non-rigourous assumption.

\begin{ass}
\label{ass:clt}
Throughout this paper, we assume sufficiently well-behaved inputs $\mathbf{x}$ and transformations such that a CLT and LLN applies to networks with independent (but not identically distributed) parameters. Specifically, 
\begin{align*}
&\phantom{{}-{}}\frac{1}{n^{(l-1)}}\sum_{i=1}^{n^{(l-1)}} \mu^{(l)}_{i} X^{(l-1)}_i - \frac{1}{n^{(l-1)}}\sum_{i=1}^{n^{(l-1)}} \mu^{(l)}_{i} \mathbb{E} [X^{(l-1)}_i ] \xrightarrow{a.s.}  0 
\end{align*}
for every $l$, and the joint distribution of 
\begin{equation*}
 \Bigg( \sum_{i=1}^{n^{(l-1)}} W^{(l)}_{ji} X^{(l-1)}_i(\mathbf{x}_1), ..., \sum_{i=1}^{n^{(l-1)}} W^{(l)}_{ji} X^{(l-1)}_i(\mathbf{x}_k) \Bigg)_{j=1}^{n^{(l)}} 
\end{equation*}
converges to a multivariate Gaussian for every finite collection of $k$ points $(\mathbf{x}_1, ..., \mathbf{x}_k)$ and layer $2 \leq  l \leq L$.
\end{ass}
While we do not claim mathematical rigour in applying this assumption, we do believe this is a mild assumption since LLNs and CLTs apply to such scaled and centered independent (but not necessarily identically distributed) random variables whose moments are well-behaved. The CLT assumption is handled rigourously by~\cite{matthews2018gaussian} in the case of independent \emph{and} identically distributed parameters. We provide empirical results in agreement with this assumption in section \S~\ref{sec:exp1}.

\subsubsection{Joint distribution of $\mathbf{X}^{(l)}$}
We now show by mathematical induction that the activation signals in the hidden layers are asymptotically jointly independent as the number of the neurons in the previous layer approaches infinity.

\textbf{Inductive basis:} It is clear that $\ranvec{X}^{(1)}$ contains jointly independent elements, since the network input is deterministic and the first layer weights contain independent rows. \textbf{Inductive Hypothesis:} Suppose the activations $\ranvec{X}^{(l-1)}$ in layer $l-1$ are jointly independent. \textbf{Inductive step}: The asymptotic distribution of pre-activations $\ranmat{W}^{(l)}\ranvec{X}^{(l-1)}$ is jointly Gaussian by Assumption~\ref{ass:clt}. Furthermore, each element is asymptotically uncorrelated, since
\begin{align*}
&\phantom{{}={}}\mathbb{E}\Big[ \lim (\ranvec{W}^{(l)}_1 \cdot \mathbf{X}^{(l-1)}) (\ranvec{W}^{(l)}_2 \cdot \mathbf{X}^{(l-1)}) \Big] \\
&= \mathbb{E}\Big[ \lim \Big(\frac{\sum_{i} \mu_i^{(l)}X^{(l-1)}_i}{n^{(l-1)}} \Big) \Big(\frac{\sum_{j} \mu_j^{(l)}X^{(l-1)}_j1}{n^{(l-1)}} \Big) \Big] \\
&= \lim \Big( \frac{\sum_{i} \mu_i^{(l)} \mathbb{E} [X^{(l-1)}_i]}{n^{(l-1)}} \Big) \Big( \frac{\sum_{j} \mu_j^{(l)} \mathbb{E} [X^{(l-1)}_j]}{n^{(l-1)}} \Big)  \\
&= \mathbb{E}\Big[ \lim (\ranvec{W}^{(l)}_1 \cdot \mathbf{X}^{(l-1)})\Big] \mathbb{E} \Big[\lim (\ranvec{W}^{(l)}_2 \cdot \mathbf{X}^{(l-1)}) \Big],
\end{align*}
and therefore independent. Therefore, $\ranvec{X}^{(l)} = \bm{\psi} (\ranmat{W}^{(l)} \ranvec{X}^{(l-1)})$ contains asymptotically jointly independent elements.

\subsubsection{Limiting conditional joint distribution of output}
By Assumption~\ref{ass:clt}, the conditional distribution of the output at any finite collection of inputs is Gaussian, and therefore the distribution of the functions at any finite number of points is governed by a GP.

\subsection{Kernels of deep networks}
\label{sec:deep_kernel_ind}
The kernel in layer $l$ is given by
\begin{align*}
&\phantom{{}={}}  \mathbb{E}\Big[ \lim_{n^{(l-1)} \to \infty} \psi( \ranvec{W}_1^{(l)} \cdot \ranvec{X}^{(l-1)}) \psi( \ranvec{W}_1^{(l)} \cdot \ranvec{X}'^{(l-1)}) \Big] \\
&= \mathbb{E} \Big[ \psi\Big(V_1 +  \mu_{av}^{(l)} \mathbb{E} \big[ X^{(l-1)}_1 \big]\Big)\psi\Big(V_2 +  \mu_{av}^{(l)} \mathbb{E} \big[ X'^{(l-1)}_1 \big]\Big) \Big] \numberthis \label{eq:kernel_l},
\end{align*}
where $\mu_{av}^{(l)} = \lim\limits_{n^{(l-1)}\to\infty}\frac{1}{n^{(l-1)}} \sum_{i=1}^{n^{(l-1)}} \mu^{(l)}_{i}$ and $(V_1, V_2)^\top = \ranvec{V}$ is a bivarate Gaussian. The covariance between $V_1$ and $V_2$ is
$$ (\sigma_{av}^{(l)})^2 \mathbb{E}\Big[ \lim_{n^{(l-2)} \to \infty} \psi( \ranvec{W}_1^{(l-1)} \cdot \ranvec{X}^{(l-2)}) \psi( \ranvec{W}_1^{(l-1)} \cdot \ranvec{X}'^{(l-2)}) \Big],$$ where $(\sigma_{av}^{(l)})^2 = \lim\limits_{n^{(l-1)}\to\infty}\frac{1}{n^{(l-1)}} \sum_{i=1}^{n^{(l-1)}} (\sigma_{i}^{(l)})^2$. This covariance is a multiple of the kernel in layer $l-1$. The variance of $V_1$ is the same as the covariance except $\ranvec{X}$ is replaced with $\ranvec{X}'$. Similarly, the variance of $V_2$ is obtained by performing the reverse replacement. This completely characterises the distribution of $\ranvec{V}$. Thus the kernel in layer $l$ is an expectation involving a zero-mean bivariate Gaussian, with a covariance matrix described by the kernel in the previous layer.

Given $\mu_{av}^{(l)}$ and $\sigma_{av}^{(l)}$, one may obtain an iterative expression for the kernel in layer $l$~\eqref{eq:kernel_l}, as in previous works. Unlike previous works, the iterative function depends not only on the kernel in layer $l-1$, but also on the first order terms $\mathbb{E} \big[ X^{(l-1)}_1 \big]$ and $\mathbb{E} \big[ X'^{(l-1)}_1 \big]$. Slightly extending the notation of~\cite{lee2017deep}, there is a deterministic function $F_\psi$ depending on the non-linearity $\psi$ and weight variances and means in layer $l$ such that
\begin{equation}
k^{(l)}_{\mathbf{x}, \mathbf{x}'\vphantom{|}} = F_\psi\Big( k^{(l-1)}_{\mathbf{x}, \mathbf{x}'\vphantom{|}}, k^{(l-1)}_{\mathbf{x}, \mathbf{x}\vphantom{|}}, k^{(l-1)}_{\mathbf{x}', \mathbf{x}'\vphantom{|}}, m^{(l-1)}_{\mathbf{x}\vphantom{|}}, m^{(l-1)}_{\mathbf{x}'\vphantom{|}}  \Big), \label{eq:ind_recurse}
\end{equation}
where $m^{(l-1)}_\mathbf{x} = \mathbb{E} \big[ X^{(l-1)}_1 \big]$ is the ``mean" in layer $l-1$. 

We derived $F^{(l)}_\psi$ exactly in the case where $\psi$ is the LReLU activation in \S~\ref{sec:single_layer_kernel}.

\section{Exchangeable priors}
In this section, we construct our generalised prior as a natural prior arising from the permutation symmetry in MLPs. Firstly, we review the notion of exchangeability. 
\subsection{Exchangeability}
\label{sec:rce_priors}
Recall that a sequence of random variables $\ranvec{Q}$ is exchangeable if $\big(Q_i)_i \stackrel{d}{=} \big( Q_{\pi(i)} \big)_i$ for all finite permutations $\pi$. That is, the joint distribution of $\ranvec{Q}$ is invariant to finitely many permutations in the indices. A generalisation of this definition to matrices $\ranmat{Q}$ called row-column exchangeability (RCE) is that $\big( Q_{ji} \big)_{ji} \stackrel{d}{=} \big( Q_{\pi_1(j) \pi_2(i)} \big)_{ji}$ for all finite permutations $\pi_1$ and $\pi_2$. 

Exchangeability generalises independence of identically distributed random variables, and the connection of these two symmetries is most elegantly described by de Finetti's theorem. De Finetti's theorem characterises the joint distribution finite subsequences of infinite exchangeable sequences as conditionally iid. De Finetti's theorem may be stated in different forms, either representing the distribution or the random variable directly. We will make use of de Finetti's theorem for RCE matrices, which says that the matrix elements are conditionally independent (but not identically distributed). Further generalisations to $n$-dimensional arrays are also available~\citep{kallenberg2006probabilistic}.
\begin{theorem}~\citep{aldous1981representations}
\label{thm:aldous}
An infinite array ${\ranmat{Q}=(Q_{ji})_{ji}}$ is RCE if and only if there exists a measurable function $F$ such that ${(Q_{ji})_{ji} \stackrel{d}{=} \big( F(A, B_j, C_i, D_{ji}) \big)_{ji},}$ where $A,$ $\ranvec{B},$ $\ranvec{C},$ and $\ranmat{D}$  are mutually iid uniform on $[0, 1]$.
\end{theorem}
We may trivially write the theorem in terms of random variables that are mutually iid uniform on $[-\sqrt3, \sqrt3]$ (having $0$ mean and unit variance). For the remainder of the paper, we will assume that $A,$ $\ranvec{B},$ $\ranvec{C},$ and $\ranmat{D}$  are mutually iid uniform on $[-\sqrt{3}, \sqrt{3}]$

\subsection{Motivating the prior}
We replace the iid weight prior distribution with a partly factorised prior of the form 
\begin{equation}
p(\ranmat{W}^{(1)}, ..., \ranmat{W}^{(L)}) = p\big(\ranmat{W}^{(1)} \big) p\big(\ranmat{W}^{(L)} \big) \prod_{l=2}^{L-1} p\big(\ranmat{W}^{(l)} \big), \label{eq:factor}
\end{equation}
allowing dependence of weights within layers in layers $2$ to $L-1$. We use independent Gaussian priors $\mathcal{N}(\ranvec{0}, \Sigma^{(1)}/n^{(0)})$ and  $\mathcal{N}\big(\ranvec{0}, \Sigma^{(L)}/n^{(L-1)}\big)$ for layers $1$ and $L$. For the marginals $p\big(\ranmat{W}^{(l)} \big), 2 \leq l \leq L-1$, we use appropriate centred and scaled RCE priors. The use of such a prior can be justified from two separate viewpoints.

\begin{enumerate}
\item The distribution over functions is invariant to permutations of the rows and columns of the weights of the MLP. Notationally, for all permutation matrices $P^{(l)}$
\begin{align*}
    \mkern-18mu &\phantom{{}={}}\ranmat{W}^{(L)} ...  \psi \big( \ranmat{W}^{(2)} \psi( \ranmat{W}^{(1)} \mathbf{x} ) \big)  \\
    \mkern-18mu &=\ranmat{W}^{(L)} \big(P^{(L-1)}\big)^T ... \psi \big( P^{(2)} \ranmat{W}^{(2)} \big(P^{(1)}\big)^T \psi(P^{(1)} \ranmat{W}^{(1)} \mathbf{x} ) \big) .
\end{align*}
We therefore constrain constant distributions inside weight equivalence classes without imposing any constraint on the function distribution:
\begin{align*}
    &\phantom{{}={}} p\big(\ranmat{W}^{(1)}, ..., \ranmat{W}^{(L)}\big) \\
    &= p\Big( P^{(1)} \ranmat{W}^{(1)} , P^{(2)} \ranmat{W}^{(2)} \big( P^{(1)}\big)^T, P^{(3)} \ranmat{W}^{(3)} \big( P^{(2)}\big)^T, ..., \ranmat{W}^{(L)} \big(P^{(L-1)}\big)^T \Big).
\end{align*}
Under the independence assumption~\eqref{eq:factor}, this implies that for $2 \leq l \leq L-1$ each $\ranmat{W}^{(l)}$ is RCE, since marginalising gives $p\big(\ranmat{W}^{(l)} \big) = p\Big(P^{(l)}\ranmat{W}^{(l)} \big(P^{(l-1)}\big)^T \Big).$ 
\item Recent work has shown that the hidden layer weight matrices of MLPs trained with a family of optimisers (including SGD and Adam) enjoy the RCE symmetry~\citep{tsuchida2018exchangeability}. Using this fact,~\cite{tsuchida2018exchangeability} study~\eqref{eq:kernel_def} for trained networks, finding closed form expressions as the input dimensionality goes to infinity for specific cases. To sample from such a prior is to sample from the distribution of solutions found by an optimiser after random initialisation, which intuitively corresponds to an informative prior.
\end{enumerate}

\subsection{Centring and scaling}
We consider an MLP with an infinite number of neurons in every hidden layer, $n^{(0)}$ inputs and $1$ output. In the hidden layers, we set the weights that are not in the upper-left $n^{(l)} \times n^{(l-1)}$ submatrix to zero. We consider a conditional distribution of functions represented by the MLP in the ordered sequence of limits $n^{(L-1)} \to \infty, ..., n^{(1)} \to \infty$ (from right to left). The internal layers have activations $\psi$ and the final layer has a linear activation. 

Motivated by Theorem~\ref{thm:aldous}, we employ the centred and scaled scheme
\begin{align*}
\label{eq:cent_scale}
\Big( W_{ji}^{(l)} \Big)_{ji} &\stackrel{d}{=} \frac{1}{\sqrt {n^{(l-1)}} } \Big( F^{(l)}_{ji} -  \mathbb{E}_{ D^{(l)}_{ji}} \big[ F^{(l)}_{ji} \big] \big( 1 - \frac{1}{\sqrt{n^{(l-1)}}} \big) \Big)_{ji} \numberthis
\end{align*}
with the short-hand $F_{ji}^{(l)} = F^{(l)}(A^{(l)},B_j^{(l)}, C_i^{(l)}, D^{(l)}_{ji})$. Note that the conditional expectation of $W^{(l)}_{ji}$ given $A^{(l)}$, $\ranvec{B}^{(l)}$ and $\ranvec{C}^{(l)}$ is $\mathbb{E}_{D_{ji}} \big[ F^{(l)}_{ji} \big]/n^{(l-1)}$, analogous to the iid case~\eqref{eq:iid_prior}. The scaling allows for applications of the CLT, and the centring allows for applications of the LLN.

We will denote $\mu_{ji}^{(l)} = \mathbb{E}_{D_{ji}} \big[ F^{(l)}_{ji} \big]$, $\bm{\mu}_j^{(l)}$ the vector whose $i$th entry is $\mu_{ji}^{(l)}$, and the diagonal matrix $\Sigma_j^{(l)}=\mathbb{E}_{D_{ji}}\Big[ (\ranvec{F}^{(l)}_{j} - \bm{\mu}_j^{(l)} \Big) \Big(\ranvec{F}^{(l)}_{j} - \bm{\mu}_j^{(l)} \Big)^T\Big]$ with $i$th diagonal entry $(\sigma_{ji}^{(l)})^2$. We will also use the centred and scaled random variable $Z_{ji}^{(l)} = (F^{(l)}_{ji} -  \mu_{ji}^{(l)}) \sigma_{ji}^{(l)}$. This leads to the more compact representation
\begin{align*}
\label{eq:cent_scale}
&\phantom{{}={}}\Big( W_{ji}^{(l)} \Big)_{ji} \stackrel{d}{=} \frac{1}{\sqrt {n^{(l-1)}} } \Big( \sigma_{ji}^{(l)} Z^{(l)}_{ji}  + \frac{\mu_{ji}^{(l)}}{\sqrt{n^{(l-1)}}} \Big)_{ji}, \numberthis
\end{align*}
which is notationally similar to the iid prior~\eqref{eq:iid_prior}. The representation~\eqref{eq:cent_scale} implies that all weights are conditionally independent (but not identically distributed) given the collection of abstract random elements
$$\ranvec{U} := \big(A^{(2)}, \ranvec{B}^{(2)}, \ranvec{C}^{(2)},  ..., A^{(L-1)}, \ranvec{B}^{(L-1)}, \ranvec{C}^{(L-1)} \big).$$ 
It is convenient to now also define $$\ranvec{U}^{\mathsf{c}}=\Big( \ranmat{W}^{(1)}, \ranmat{D}^{(2)}, ..., \ranmat{D}^{(L-1)}, \ranvec{W}^{(L)} \Big),$$
i.e. $\ranvec{U}^{\mathsf{c}}$ is the collection of random variables that are not part of $\ranvec{U}$.

After conditioning on $\ranvec{U}$, the only difference between the present and the independent case~\eqref{eq:iid_prior} is that the means and variances vary not only in the columns $i$, but also in the rows $j$. This does not pose a problem in terms of showing convergence to a GP, but does make it difficult to evaluate the kernel of the network analytically. In order to handle this issue, we will later restrict our attention to a large subset of all possible $F^{(l)}$'s.

\subsection{Convergence of conditional distribution to a GP}
We firstly consider the conditional distribution of functions given $\ranvec{U}$ in the infinite width limit. 
\subsubsection{Limiting conditional joint distribution of $\mathbf{X}^{(l)}$}
\textbf{Inductive basis:} It is clear that $\ranvec{X}^{(1)}$ are conditionally jointly independent given $\ranvec{U}$ (since in the first layer we retained an independent Gaussian prior). \textbf{Inductive Hypothesis:} Suppose the activations $\ranvec{X}^{(l-1)}$ in layer $l-1$ are conditionally jointly independent given $\ranvec{U}$. \textbf{Inductive step}: The asymptotic conditional distribution of pre-activations $\ranmat{W}^{(l)}\ranvec{X}^{(l-1)}$ given $\ranvec{U}$ is jointly Gaussian by Assumption~\ref{ass:clt}. Additionally, each element is conditionally asymptotically uncorrelated, since
\begin{align*}
&\phantom{{}={}}\mathbb{E}_{\ranvec{U}^{\mathsf{c}}} \Big[ \lim (\ranvec{W}^{(l)}_1 \cdot \mathbf{X}^{(l-1)}) (\ranvec{W}^{(l)}_2 \cdot \mathbf{X}^{(l-1)}) \Big] \\
&= \mathbb{E}_{\ranvec{U}^{\mathsf{c}}} \Big[ \lim \Big(\frac{\sum_{i} \mu_{1i}^{(l)}X^{(l-1)}_i}{n^{(l-1)}} \Big) \Big(\frac{\sum_{j} \mu_{2j}^{(l)}X^{(l-1)}_j}{n^{(l-1)}} \Big) \Big] \\
&= \mathbb{E}_{\ranvec{U}^{\mathsf{c}}} \Bigg[ \lim \frac{\sum_{i} \mu_{1i}^{(l)}X^{(l-1)}_i}{n^{(l-1)}} \Bigg] \mathbb{E}_{\ranvec{U}^{\mathsf{c}}} \Bigg[ \lim \frac{\sum_{j} \mu_{2j}^{(l)}X^{(l-1)}_j}{n^{(l-1)}}  \Bigg] \\
&=\mathbb{E}_{\ranvec{U}^{\mathsf{c}}} \Big[ \lim (\ranvec{W}^{(l)}_1 \cdot \mathbf{X}^{(l-1)}) \Big] \mathbb{E}_{\ranvec{U}^{\mathsf{c}}} \Big[ \lim (\ranvec{W}^{(l)}_2 \cdot \mathbf{X}^{(l-1)}) \Big],
\end{align*}
and therefore independent. In order to go from the second to the third line, we used the fact that the integrands are conditionally deterministic given $\ranvec{U}$, so the expectation factors. Therefore, $\ranvec{X}^{(l)} = \bm{\psi} (\ranmat{W}^{(l)} \ranvec{X}^{(l-1)})$ is conditionally jointly independent given $\ranvec{U}$.

\subsubsection{Limiting conditional joint distribution of output}
By Assumption~\ref{ass:clt}, the conditional distribution of the output at any finite collection of inputs is Gaussian, and therefore the distribution in function space is governed by a GP.

\subsection{Kernels of deep networks}
\label{sec:deep_kernel_rce}
The kernel in layer $l$ is given by
\begin{align*}
&\phantom{{}={}}\lim_{n^{(l)} \to \infty} \frac{1}{n^{(l)}} \sum_{j=1}^{n^{(l)}}  \mathbb{E}_{\ranvec{U}^{\mathsf{c}}} \Big[ \lim_{n^{(l-1)} \to \infty} \psi( \ranvec{W}_j^{(l)} \cdot \ranvec{X}^{(l-1)}) \psi( \ranvec{W}_j^{(l)} \cdot \ranvec{X}'^{(l-1)}) \Big] \\
&= \mathbb{E}_{B^{(l)}_1, \ranvec{U}^{\mathsf{c}}} \Big[ \lim_{n^{(l-1)} \to \infty} \psi( \ranvec{W}_1^{(l)} \cdot \ranvec{X}^{(l-1)}) \psi( \ranvec{W}_1^{(l)} \cdot \ranvec{X}'^{(l-1)}) \Big].
\end{align*}
The deterministic component of $\ranvec{W}_1^{(l)} \cdot \ranvec{X}^{(l-1)}$ due to the mean is almost surely
\begin{align*}
&\phantom{{}={}}\lim\limits_{n^{(l-1)}\to\infty}\frac{1}{n^{(l-1)}} \sum_{i=1}^{n^{(l-1)}} \mu^{(l)}_{1i} X^{(l-1)}_i \\
&= \mathbb{E}_{C_1^{(l)}, B^{(l-1)}_1, \ranvec{U}^{\mathsf{c}}} \big[ \mu^{(l)}_{11} X^{(l-1)}_1 \big] \\
&=  \mathbb{E}_{C_1^{(l)}} \Big[ \mu^{(l)}_{11} \Big] \mathbb{E}_{B^{(l-1)}_1, \ranvec{U}^{\mathsf{c}}} \big[ X^{(l-1)}_1 \big] \numberthis \label{eq:first_order_rce}
\end{align*}
by Assumption~\ref{ass:clt}. The random component $\ranvec{V}=(V_1, V_2)^\top$ is a $0$-mean bivarate Gaussian. The conditional covariance between $V_1$ and $V_2$ is 
\begin{align*}
&\phantom{{}={}} \lim\limits_{n^{(l-1)}\to\infty}\frac{1}{n^{(l-1)}} \sum_{i=1}^{n^{(l-1)}} (\sigma_{1i}^{(l)})^2 \mathbb{E}_{\ranvec{U}^{\mathsf{c}}}\Big[ \lim_{n^{(l-2)} \to \infty} \psi( \ranvec{W}_i^{(l-1)} \cdot \ranvec{X}^{(l-2)}) \psi( \ranvec{W}_i^{(l-1)} \cdot \ranvec{X}'^{(l-2)}) \Big] \\
&= \mathbb{E}_{C_1^{(l)}} \big[(\sigma^{(l)}_{11})^2 \big] \mathbb{E}_{B^{(l-1)}_1, \ranvec{U}^{\mathsf{c}}} \Big[ \lim_{n^{(l-2)} \to \infty} \psi( \ranvec{W}_1^{(l-1)} \cdot \ranvec{X}^{(l-2)}) \psi( \ranvec{W}_1^{(l-1)} \cdot \ranvec{X}'^{(l-2)}) \Big],  \numberthis \label{eq:second_order_rce}
\end{align*}
which is the kernel in layer $l-1$ with an additional factor. The variance of $V_1$ is the same as the covariance except $\ranvec{X}$ is replaced with $\ranvec{X}'$. Similarly, the variance of $V_2$ is obtained by performing the reverse replacement. This completely characterises the distribution of $\ranvec{V}$.

Unfortunately, evaluating the kernel in the same manner as \S~\ref{sec:single_layer_kernel} is made difficult by the fact that we are now required to marginalise out $B^{(l-1)}_1$ in the second and first order terms~\eqref{eq:first_order_rce} and~\eqref{eq:second_order_rce}. We could approximate the integration using Monte Carlo methods, but for this present study we instead prefer to make a simplifying assumption. In order to make this integration tractable, we further restrict our attention to a subclass of $F$'s for which dependence on $B$ may be factored out in the following sense.

\begin{ass}
\label{ass:same_means}
For all $l$, there exist some $G^{(l)}: \mathbb{R} \to \mathbb{R}$ and $H^{(l)}: \mathbb{R}^3 \to \mathbb{R}$ such that
\begin{align*}
F^{(l)}(A,B,C,D) = G^{(l)} (B) \, H^{(l)}(A,C,D).
\end{align*}
\end{ass}
By absolute homogeneity~\eqref{eq:abs_homo}, this allows one to obtain an iterative expression for the kernel in layer $l$ since the conditional covariance between $V_1$ and $V_2$,~\eqref{eq:second_order_rce} is
\begin{align*}
&= \mathbb{E}_{C_1^{(l)}} \big[(\sigma^{(l)}_{11})^2 \big] \mathbb{E}_{B^{(l-1)}_1, \ranvec{U}^{\mathsf{c}}} \Big[ \lim_{n^{(l-2)} \to \infty} \psi( \ranvec{W}_1^{(l-1)} \cdot \ranvec{X}^{(l-2)}) \psi( \ranvec{W}_1^{(l-1)} \cdot \ranvec{X}'^{(l-2)}) \Big] \\
&= \mathbb{E}_{C_1^{(l)}} \big[(\sigma^{(l)}_{11})^2 \big] \mathbb{E}_{B^{(l-1)}_1}\Big[\big(  G^{(l-1)}(B_1) \big)^2\Big] \mathbb{E}_{\ranvec{U}^{\mathsf{c}}} \Big[ \lim_{n^{(l-2)} \to \infty} \psi( \ranvec{T}_1^{(l-1)} \cdot \ranvec{X}^{(l-2)}) \psi( \ranvec{T}_1^{(l-1)} \cdot \ranvec{X}'^{(l-2)}) \Big],
\end{align*}
where $T_{\cdot, \cdot}^{(\cdot)} = W_{\cdot, \cdot}^{(\cdot)}/G^{(\cdot)}(B_{\cdot})$ are the weights with $H$ taking the role of $F$. A similar calculation holds for the first order terms~\eqref{eq:first_order_rce}, resulting in a factor of $\mathbb{E}_{B^{(l-1)}_1}\Big[\big|  G^{(l-1)}(B_1) \big|\Big]$.

As in the independent parameter case~\eqref{eq:ind_recurse}, in the notation of~\cite{lee2017deep}, there is a deterministic function $F_\psi$ depending on the non-linearity $\psi$ such that
\begin{equation}
k^{(l)}_{\mathbf{x}, \mathbf{x}'\vphantom{|}} = F_\psi\Big( k^{(l-1)}_{\mathbf{x}, \mathbf{x}'\vphantom{|}}, k^{(l-1)}_{\mathbf{x}, \mathbf{x}\vphantom{|}}, k^{(l-1)}_{\mathbf{x}', \mathbf{x}'\vphantom{|}}, m^{(l-1)}_{\mathbf{x}\vphantom{|}}, m^{(l-1)}_{\mathbf{x}'\vphantom{|}}  \Big),
\end{equation}
where $m^{(l-1)}_\mathbf{x} = \mathbb{E}_{\ranvec{U}^{\mathsf{c}}} \big[ X^{(l-1)}_1 \big]$ is the ``mean" in layer $l-1$. Furthermore, we have derived $F_\psi$ exactly in the case where $\psi$ is the LReLU activation, as  in \S~\ref{sec:single_layer_kernel}.

\section{The limiting model}
The only elements inside $\ranvec{U}$ that appear in the kernel in the last layer $K^{(L-1)}_{\ranvec{U}}$ are the $A^{(l)}$. The dependence on the $A^{(l)}$ is only through the abstract objects $\bm{\sigma} = \big( \sigma^{(l)} \big)_{l=2}^{L-1}$ and $\bm{\mu} = \big( \mu^{(l)} \big)_{l=2}^{L-1}$, which may be interpreted as hyperparameters of the kernel.

Conditional on $\ranvec{U}$, the posterior predictive distribution is
\begin{align*}
\mathbf{f}_* &| \ranvec{U}, X_*, X, \ranvec{y} \sim \mathcal{N}\big( \text{mean}(X_*), \text{cov}(X_*, X_*)  \big), \quad \text{where} \numberthis \label{eq:cond_pred} \\
\text{mean}(X_*) &= K^{(L-1)}_{\ranvec{U}}(X_*, X)  \big(K^{(L-1)}_{\ranvec{U}}(X, X)+s^2 I\big)^{-1} \mathbf{y}, \quad \text{and} \\
\text{cov}(X_*, X_*) &= K^{(L-1)}_{\ranvec{U}}(X_*, X_*) -  K^{(L-1)}_{\ranvec{U}}(X_*, X)\big( K^{(L-1)}_{\ranvec{U}}(X, X)+s^2I\big)^{-1} K^{(L-1)}_{\ranvec{U}}(X, X_*).
\end{align*}

We can make predictions by sampling from or using the mean of
\begin{align*}
&\phantom{{}={}}p\big(\mathbf{f}_*  | X_*, X, \mathbf{y} \big) \\
&= \int p\big( \mathbf{f}_* | \ranvec{U}, X_*, X, \mathbf{y} \big) p\big(\ranvec{U} | X_*, X, \mathbf{y} \big) \, d\ranvec{U} \\
&= \int p\big( \mathbf{f}_* | \bm{\mu},\bm{\sigma}, X_*, X, \mathbf{y} \big) p\big(\bm{\mu},\bm{\sigma} | X_*, X, \mathbf{y} \big) \, d\bm{\mu}d\bm{\sigma}. \numberthis \label{eq:predictive}
\end{align*}
This corresponds to a ``fully" Bayesian treatment of GPs, where we marginalise out the posterior distribution over hyperparameters. We ignore the third level of inference where one posits a prior over models, as unlike the hyperprior, this did not arise naturally in our construction.

An interesting special case of~\eqref{eq:predictive} is when the dependence on all the $A^{(l)}$ is removed from the representation of Theorem~\ref{thm:aldous} in~\eqref{eq:cent_scale}. This corresponds to what is called a dissociated RCE (dRCE) array~\citep{aldous1981representations}. To get a sense for what it means for an RCE array to be dRCE, it follows from the representation Theorem~\ref{thm:aldous} that in an infinite dRCE array, any two groups of elements with disjoint row and column index sets are independent. When the weight matrices are dRCE, it is immediate that $\bm{\mu}$ and $\bm{\sigma}$ are deterministic. Thus, 
\begin{rmk}
When $\big( F^{(l)}_{ji} \big)_{ji}$ in~\eqref{eq:cent_scale} is dRCE for each $l$, the hyper-posterior reduces to a point mass and the limiting model corresponds to a GP with point hyperparameters.
\end{rmk}
Note that the dissociated property is a sufficient but not necessary condition for the limiting model to have point hyperparameters. For example, $F3$  in Table~\ref{tab:fs} is not dRCE but does result in a limiting GP with a fixed hyperparameter.

\subsection{Hyperparameter intuition}
\subsubsection{Reparameterisation}
\label{sec:reparam}
Using the absolute homogeneity~\eqref{eq:abs_homo} of the LReLU, a reparameterisation allows for more easily interpretable hyperparameters. We may factor out the standard deviations of the weights in each layer. Letting $\ranmat{W}^{(l)} = \frac{1}{\sqrt{n^{(l-1)}}} \big( \sigma^{(l)} \ranmat{Z}^{(l)} + \frac{\mu^{(l)}}{\sqrt{n^{(l-1)}}} \big)$ where $\ranmat{Z}^{(l)}$ has variance $1$ and mean $0$, the mapping performed by a single layer can be written as
\begin{align*}
\frac{1}{\sqrt{n^{(l-1)}}} \sigma^{(l)} \ranvec{\psi} \Big( \big( \ranmat{Z}^{(l)} + \frac{\mu^{(l)}}{\sigma^{(l)}\sqrt{n^{(l-1)}}} \big) \cdot \ranvec{X}^{(l-1)}\Big).
\end{align*}
We define the signal to noise ratio
\begin{equation}
d^{(l)}:= \mu^{(l)}/\sigma^{(l)}. \label{eq:reparam}
\end{equation}
Under the reparametersiation~\eqref{eq:reparam}, it is clear that $d^{(l)}$ controls the shape of the sample functions in the GP, and $\sigma^{(l)}$ controls the scale.

\subsubsection{Marginal likelihood surface in deep networks}
In a deep network, if $\sigma^2$ is very different to $2/(1+a^2)$, the signal will either vanish or explode as it passes through the network, resulting in very small or large outputs. Therefore, a reasonable value of $\sigma^2$ along $\mu=0$ is $2$ when the LReLU negative gradient parameter is $a=0$ and $2/(1+a^2)$ for general $a$, as in~\eqref{eq:arc-cosine}. As $\mu$ becomes more negative, the signal before the activation function is applied in intermediate layers is more negative, resulting in a post-activation signal with small magnitude. Therefore, we expect a more negative $\mu$ to be associated with a larger $\sigma^2$ to amplify the post-activation signal. 

As a general rule of thumb for increasingly deep networks, we expect a maximum along $\mu=0$ at $\sigma^2 \approx 2$, and a general trend of $\mu$ becoming more negative as $\sigma^2$ increases.

\section{Experiments}
We are interested in using our model as a conceptual tool to better understand inductive biases in MLPs rather than achieve state of the art performance in machine learning tasks. With this in mind, we conduct experiments with the aim of investigating three broad empirical questions:
\begin{enumerate}
\item Does the prior of finite width networks closely imitate the limiting GP model?
\item In the point hyperparameter case, do the additional hyperparameters of the kernel afforded by the use of weights with non-zero means give any benefit above zero means?
\item Does marginalisation of hyperparameters enjoy any advantages over using a point hyperparameter model?
\end{enumerate}

For all experiments, we used the same $\mu^{(l)}=\mu$ and $\sigma^{(l)}=\sigma$ in each layer.

\subsection{Datasets}
The experimental results for Questions $2$ and $3$ concern three datasets.
\begin{enumerate}
\item \texttt{Sine}, a one dimensional regression problem $y = \sin(x) + \epsilon$, where $\epsilon ~\sim \mathcal{N}(0, 0.1)$. We take $10$ training examples on a uniform grid over $[-\sqrt{3}, \sqrt{3}]$ and $100$ test examples sampled uniformly over the same interval.
\item \texttt{Smooth XOR}, a two dimensional regression problem $y = -x_1 x_2 \exp(2-x_1^2-x_2^2) + \epsilon$, where $\epsilon ~\sim \mathcal{N}(0, 0.1)$. We take every permutation of $(\pm 1, \pm 1)$ as the training set containing $4$ examples, and $100$ uniformly sampled points on $[-2, 2]$ as the testing set.
\item \texttt{Snelson}~\citep{snelson2006sparse}, a one dimensional synthetic regression problem in which we take the original training set of $200$ one dimensional regression pairs, and split the data into $10$ equally spaced training pairs and use the remaining $190$ points for testing.
\iffalse
\item \textcolor{red}{\texttt{Mauna}~\citep{keeling1995interannual}, a one dimensional real-world dataset containg years and months as inputs and the average CO2 emissions at the Mauna volcano as targets. The data exhibits periodic behaviour superimposed on a longer trend, for which our current priors expressed through LReLU activations are not suitable. A periodic model would be more appropriate~\citep{pearce2019expressive}, but we include this dataset to study the behaviour of the model applied to unsuitable data.}
\item \textcolor{red}{\texttt{Yacht}~\citep{Dua:2019}, a six dimensional real-world dataset containing dimensionless numbers as inputs and a quantity related to the hydrodynamic performance of yachts as targets. We randomly split the data into $100$ training instances and use the remaining $208$ points for testing.}
\fi
\end{enumerate}
For each dataset, we performed GP regression with a Gaussian likelihood with fixed noise variance $0.1$.

\subsection{Question 1}
\label{sec:exp1}
\begin{table*}[!t]
\begin{tabularx}{\linewidth}{l|l|l}
    $\bm{F(A,B,C,D)}$ & \textbf{MLP \newline weights} & \textbf{GP hyperparameters $\bm{(\mu, \sigma)}$} \\
                  \hline
    $F1=\sqrt{2}D$ & Centred, scaled iid uniform & $(0,\sqrt{2})$ \\
    $F2=2\sqrt{2}D-0.5$ & iid uniform & $(-0.5, \sqrt{8})$\\
    $F3=\sqrt{2}D-1.5AC$ & RCE  & $(0, \sqrt{2})$ \\
    $F4=\sqrt{2}D(A+\sqrt{3})-$ & RCE  & $\Big(-0.1A^2-0.4, 2\big|A+\sqrt{3}\big| \Big)$ \\ $ \qquad \phantom{=} 0.1A^2C^2 - 0.4$ & & 
\end{tabularx}
\caption{Description of different weight sampling schemes.}
\label{tab:fs}
\end{table*}

\begin{figure*}[!t]
\centering
\begin{minipage}[t]{.3\linewidth}
\includegraphics[width=\linewidth]{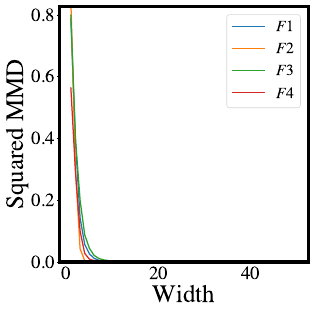}
\subcaption{$L=4$}
\end{minipage} 
\begin{minipage}[t]{.3\linewidth}
\includegraphics[width=\linewidth]{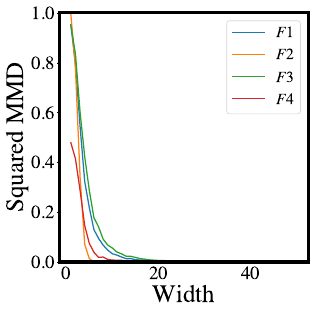}
\subcaption{$L=8$}
\end{minipage}
\begin{minipage}[t]{.3\linewidth}
\includegraphics[width=\linewidth]{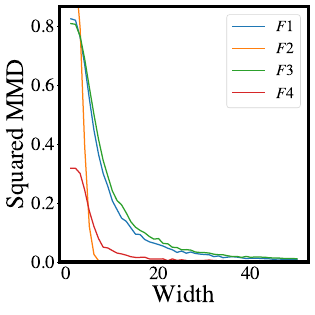}
\subcaption{$L=16$} 
\end{minipage} \\
\begin{minipage}[t]{.3\linewidth}
\includegraphics[width=\linewidth]{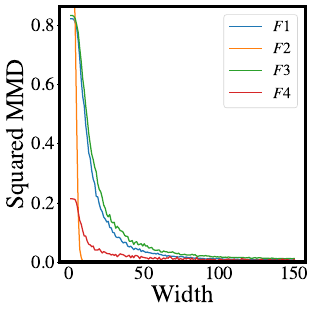}
\subcaption{$L=32$}
\end{minipage}
\begin{minipage}[t]{.3\linewidth}
\includegraphics[width=\linewidth]{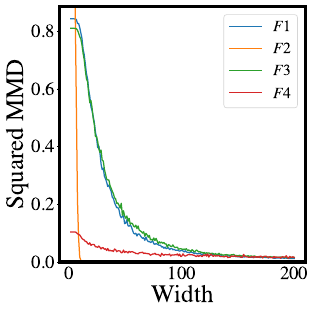}
\subcaption{$L=64$}
\end{minipage}
\caption{Unbiased estimate of the squared MMD between $f_{MLP}$ and $f_{GP}$ for different random weight schemes and network depths.}
\label{fig:mmd}
\end{figure*}

We define a random MLP $f_{MLP}: \mathbb{R}^{10} \to \mathbb{R}$ of finite hidden layer width and a potentially \emph{random} (with random kernel hyperparameters) GP with samples $f_{GP}: \mathbb{R}^{10} \to \mathbb{R}$. We measure the empirical Maximum Mean Discrepency (MMD) as a function of hidden layer width between $f_{MLP}$ and $f_{GP}$ over $4$ randomly chosen points in input space $\ranmat{S} \sim \mathcal{N}(\mathbf{0}, I_{40})$. We sample the function values at these points $2000$ times for calculation of an unbiased estimate of the MMD. We use the kernel $\kappa_{ij} = \exp \big( -\Vert f_{MLP}(\ranvec{S}_i) - f_{GP}(\ranvec{S}_j) \Vert^2 \big)$ for the MMD, without regard for the most suitable choice of hyperparameters.

For simplicity, we use the same $F^{(l)}_{ji} \equiv F$ in every layer. To better cover the space of hyperparameters, we try different values of $L$ and four different forms $F$ of differing levels of ``complexity", as described in Table~\ref{tab:fs}. The plots are shown in Figure~\ref{fig:mmd}. In each of the plots, the MMD decreases to $0$ as the width increases. As in~\cite{matthews2018gaussian}, we observe that the deeper models converge more slowly.

In addition, the choice of $F$ clearly changes the speed of convergence. $F1$ and $F3$ both result in ``vanilla" GP's with previously derived kernels~\citep{NIPS2009_3628}. The scaling is chosen such that the norm of the signal stays constant throughout the layers, as is approximately achieved using He initialisation in finitely wide neural networks~\citep{he2015delving}. It is worth remarking that $F_3$ has dependent parameters, even though it still results in the same limiting model.

The negative mean in $F2$ would result in the norm of the signals \emph{decreasing} if we were to use the same $\sqrt2$ scaling in $D$, since the signals inside the network are rectified. We therefore scale the variance up to avoid a vanishing signal in deeper layers. For variety, in $F4$ we made the means and variances depend on $A$.

\begin{figure*}[!t]
\centering
\begin{minipage}[t]{.15\linewidth}
\includegraphics[width=\linewidth]{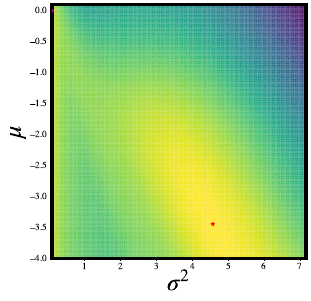}
\subcaption{$L=2$}
\end{minipage}\hfill
\begin{minipage}[t]{.15\linewidth}
\includegraphics[width=\linewidth]{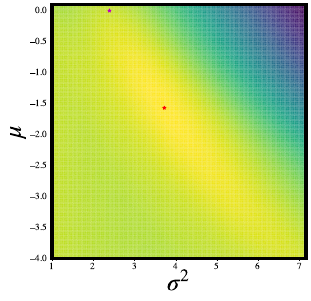}
\subcaption{$L=4$}
\end{minipage}\hfill
\begin{minipage}[t]{.15\linewidth}
\includegraphics[width=\linewidth]{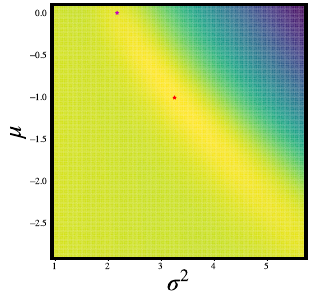}
\subcaption{$L=8$}
\end{minipage}\hfill
\begin{minipage}[t]{.15\linewidth}
\includegraphics[width=\linewidth]{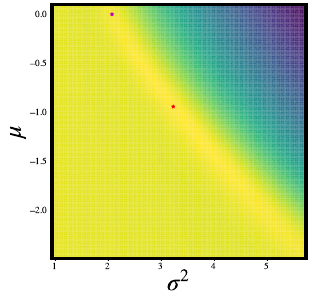}
\subcaption{$L=16$}
\end{minipage}\hfill
\begin{minipage}[t]{.15\linewidth}
\includegraphics[width=\linewidth]{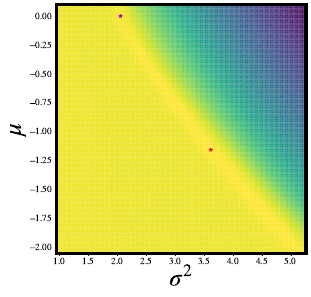}
\subcaption{$L=32$}
\end{minipage}\hfill
\begin{minipage}[t]{.15\linewidth}
\includegraphics[width=\linewidth]{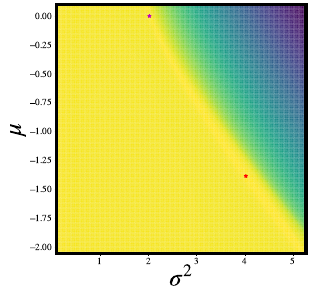}
\subcaption{$L=64$}
\end{minipage}\hfill
\begin{minipage}[t]{.15\linewidth}
\includegraphics[width=\linewidth]{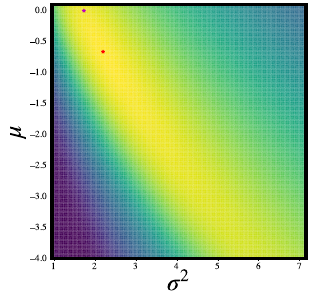}
\subcaption{$L=2$}
\end{minipage}\hfill
\begin{minipage}[t]{.15\linewidth}
\includegraphics[width=\linewidth]{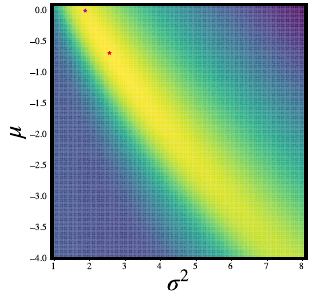}
\subcaption{$L=4$}
\end{minipage}\hfill
\begin{minipage}[t]{.15\linewidth}
\includegraphics[width=\linewidth]{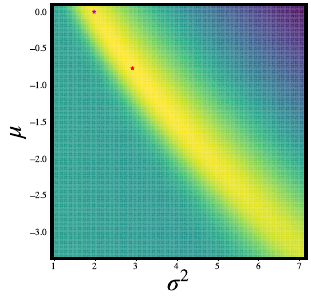}
\subcaption{$L=8$}
\end{minipage}\hfill
\begin{minipage}[t]{.15\linewidth}
\includegraphics[width=\linewidth]{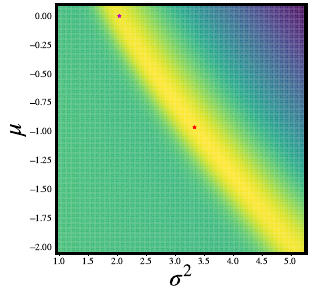}
\subcaption{$L=16$}
\end{minipage}\hfill
\begin{minipage}[t]{.15\linewidth}
\includegraphics[width=\linewidth]{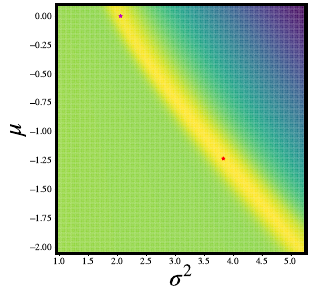}
\subcaption{$L=32$}
\end{minipage}\hfill
\begin{minipage}[t]{.15\linewidth}
\includegraphics[width=\linewidth]{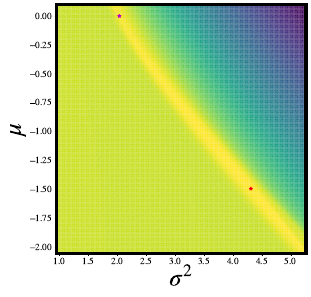}
\subcaption{$L=64$}
\end{minipage}\hfill
\begin{minipage}[t]{.15\linewidth}
\includegraphics[width=\linewidth]{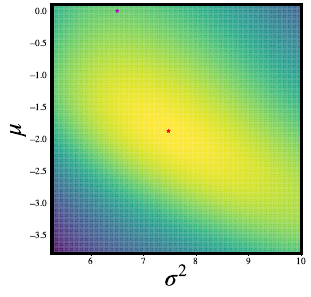}
\subcaption{$L=2$}
\end{minipage}\hfill
\begin{minipage}[t]{.15\linewidth}
\includegraphics[width=\linewidth]{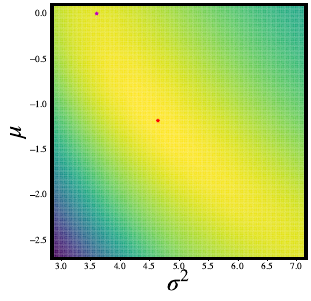}
\subcaption{$L=4$}
\end{minipage}\hfill
\begin{minipage}[t]{.15\linewidth}
\includegraphics[width=\linewidth]{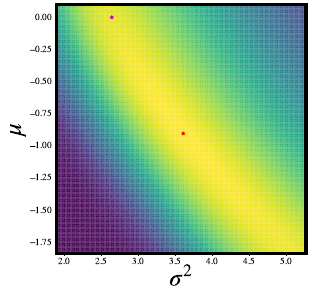}
\subcaption{$L=8$}
\end{minipage}\hfill
\begin{minipage}[t]{.15\linewidth}
\includegraphics[width=\linewidth]{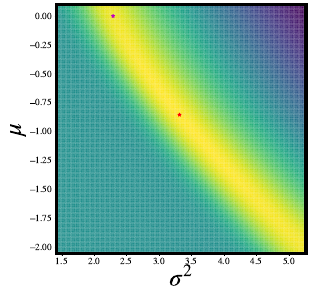}
\subcaption{$L=16$}
\end{minipage}\hfill
\begin{minipage}[t]{.15\linewidth}
\includegraphics[width=\linewidth]{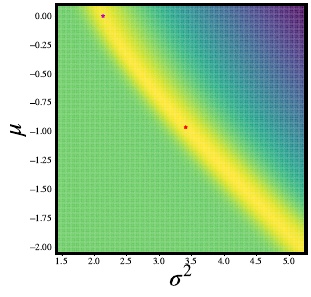}
\subcaption{$L=32$}
\end{minipage}\hfill
\begin{minipage}[t]{.15\linewidth}
\includegraphics[width=\linewidth]{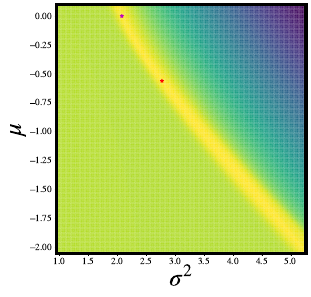}
\subcaption{$L=64$}
\end{minipage}\hfill
\caption{Log marginal likelihood. The red points indicate location of maxima, and the magenta points indicate the location of the maxima constrained along $\mu=0$ (up to a discrete grid of $200 \times 200$). (a - f) \texttt{Sine}. (g - f) \texttt{Smooth XOR}. (m - r) \texttt{Snelson}.}
\label{fig:ll}
\end{figure*}

\begin{figure*}[!t]
\centering
\begin{minipage}[t]{.15\linewidth}
\includegraphics[width=\linewidth]{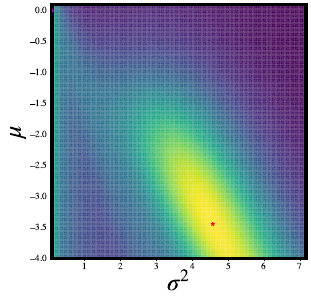}
\subcaption{$L=2$}
\end{minipage}\hfill
\begin{minipage}[t]{.15\linewidth}
\includegraphics[width=\linewidth]{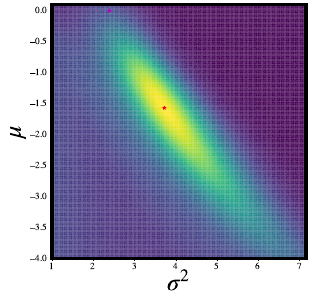}
\subcaption{$L=4$}
\end{minipage}\hfill
\begin{minipage}[t]{.15\linewidth}
\includegraphics[width=\linewidth]{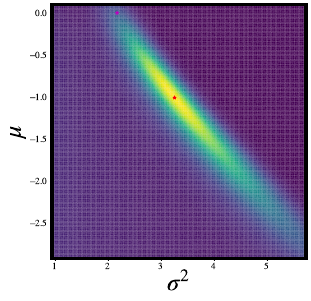}
\subcaption{$L=8$}
\end{minipage}\hfill
\begin{minipage}[t]{.15\linewidth}
\includegraphics[width=\linewidth]{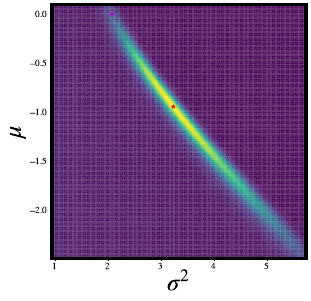}
\subcaption{$L=16$}
\end{minipage}\hfill
\begin{minipage}[t]{.15\linewidth}
\includegraphics[width=\linewidth]{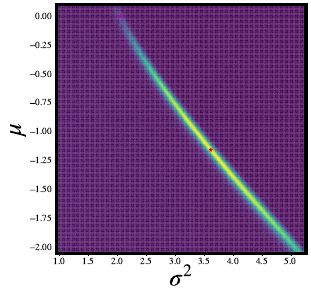}
\subcaption{$L=32$}
\end{minipage}\hfill
\begin{minipage}[t]{.15\linewidth}
\includegraphics[width=\linewidth]{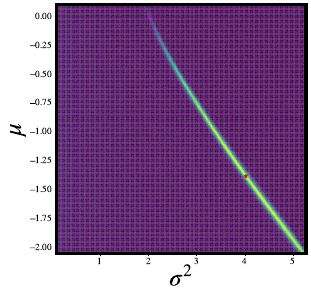}
\subcaption{$L=64$}
\end{minipage}\hfill
\begin{minipage}[t]{.15\linewidth}
\includegraphics[width=\linewidth]{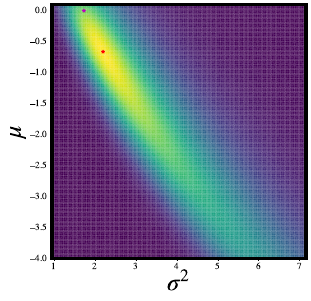}
\subcaption{$L=2$}
\end{minipage}\hfill
\begin{minipage}[t]{.15\linewidth}
\includegraphics[width=\linewidth]{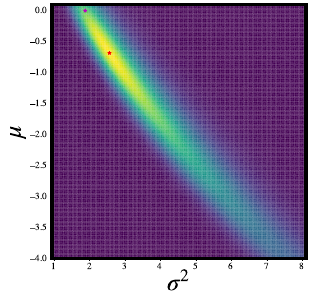}
\subcaption{$L=4$}
\end{minipage}\hfill
\begin{minipage}[t]{.15\linewidth}
\includegraphics[width=\linewidth]{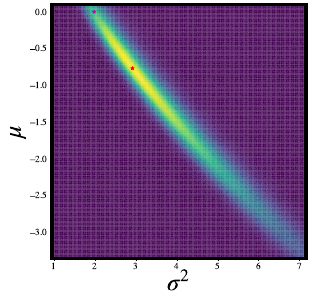}
\subcaption{$L=8$}
\end{minipage}\hfill
\begin{minipage}[t]{.15\linewidth}
\includegraphics[width=\linewidth]{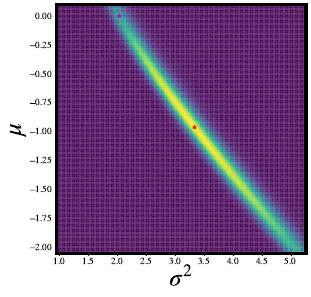}
\subcaption{$L=16$}
\end{minipage}\hfill
\begin{minipage}[t]{.15\linewidth}
\includegraphics[width=\linewidth]{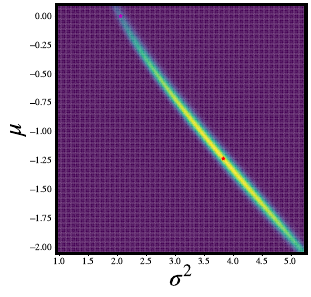}
\subcaption{$L=32$}
\end{minipage}\hfill
\begin{minipage}[t]{.15\linewidth}
\includegraphics[width=\linewidth]{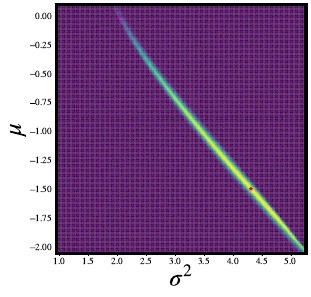}
\subcaption{$L=64$}
\end{minipage}\hfill
\begin{minipage}[t]{.15\linewidth}
\includegraphics[width=\linewidth]{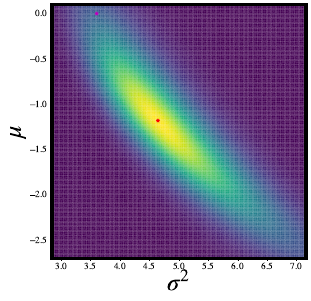}
\subcaption{$L=2$}
\end{minipage}\hfill
\begin{minipage}[t]{.15\linewidth}
\includegraphics[width=\linewidth]{low_res_figures/likelihoods/snelson/likelihood4_False-1.png}
\subcaption{$L=4$}
\end{minipage}\hfill
\begin{minipage}[t]{.15\linewidth}
\includegraphics[width=\linewidth]{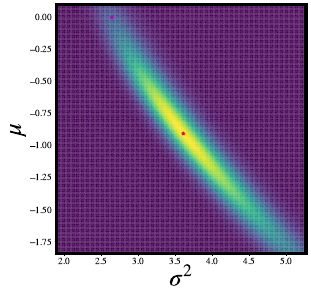}
\subcaption{$L=8$}
\end{minipage}\hfill
\begin{minipage}[t]{.15\linewidth}
\includegraphics[width=\linewidth]{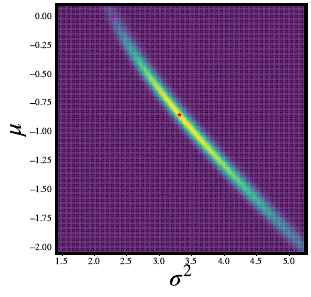}
\subcaption{$L=16$}
\end{minipage}\hfill
\begin{minipage}[t]{.15\linewidth}
\includegraphics[width=\linewidth]{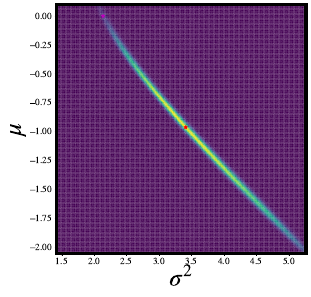}
\subcaption{$L=32$}
\end{minipage}\hfill
\begin{minipage}[t]{.15\linewidth}
\includegraphics[width=\linewidth]{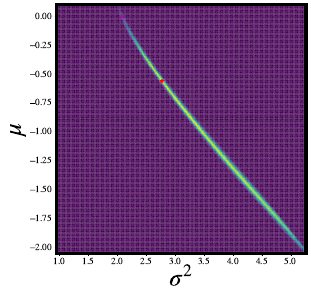}
\subcaption{$L=64$}
\end{minipage}\hfill
\caption{Marginal likelihood. The red points indicate location of maxima, and the magenta points indicate the location of the maxima constrained along $\mu=0$ (up to a discrete grid of $200 \times 200$). (a - f) \texttt{Sine}. (g - f) \texttt{Smooth XOR}. (m - r) \texttt{Snelson}.}
\label{fig:marginal_likelihood}
\end{figure*}

\subsection{Question 2}
To answer Question 2, we demonstrate firstly that the model evidence (marginal likelihood) obtains its maximum at a value $\mu \neq 0$, and secondly that the test error obtained by the best model with $\mu\neq0$ can be lower than the test error obtained by the best model with $\mu = 0$.

\begin{figure*}[!t]
\centering
\begin{minipage}[t]{.3\linewidth}
\includegraphics[width=\linewidth]{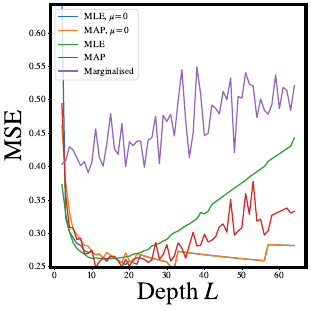}
\subcaption{}
\end{minipage}\hfill
\begin{minipage}[t]{.3\linewidth}
\includegraphics[width=\linewidth]{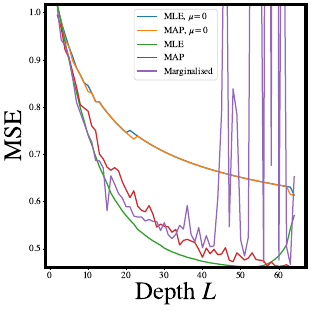}
\subcaption{}
\end{minipage}\hfill
\begin{minipage}[t]{.3\linewidth}
\includegraphics[width=\linewidth]{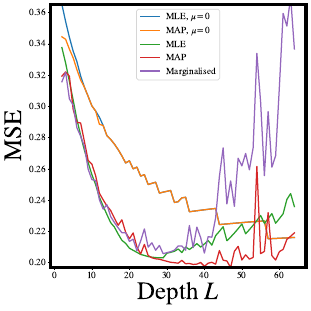}
\subcaption{}
\end{minipage} 
\caption{Performance measures as a function of depth. (a, d) \texttt{Sine}. (b, e) \texttt{Smooth XOR}. (c, f) \texttt{Snelson}. }
\label{fig:performance}
\end{figure*}

To this effect, in Figures~\ref{fig:ll} and~\ref{fig:marginal_likelihood} we plot the log marginal likelihood and marginal likelihood in the $(\mu, \sigma^2)$ plane (where $\mu^{(l)}=\mu$, $\sigma^{(l)}=\sigma$ is the same in each layer) for networks of different depths. We also plot the test mean squared error (MSE) to the mean of the posterior predictive distribution as a function of depth for models with optimal hyperparameters with respect to the marginal likelihood (MLE) for models with $\mu =0$ and $\mu \neq 0$ in Figure~\ref{fig:performance}. 

The maximum likelihood occurs at $\mu \neq 0$, for every depth and problem, and for some depths and problems, a lower test error is obtained by models with $\mu \neq 0$.

\subsection{Question 3}
\begin{figure*}[!htb]
\centering
\begin{minipage}[t]{.15\linewidth}
\includegraphics[width=\linewidth]{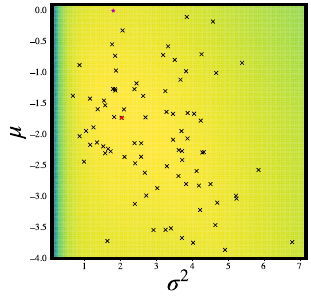}
\subcaption{$L=2$}
\end{minipage}\hfill
\begin{minipage}[t]{.15\linewidth}
\includegraphics[width=\linewidth]{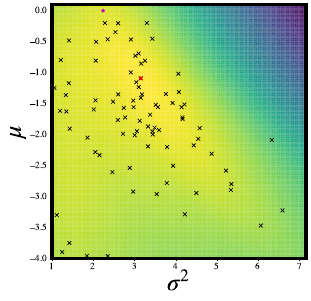}
\subcaption{$L=4$}
\end{minipage}\hfill
\begin{minipage}[t]{.15\linewidth}
\includegraphics[width=\linewidth]{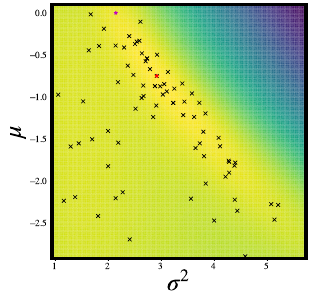}
\subcaption{$L=8$}
\end{minipage}\hfill
\begin{minipage}[t]{.15\linewidth}
\includegraphics[width=\linewidth]{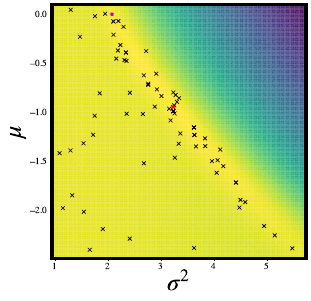}
\subcaption{$L=16$}
\end{minipage}\hfill
\begin{minipage}[t]{.15\linewidth}
\includegraphics[width=\linewidth]{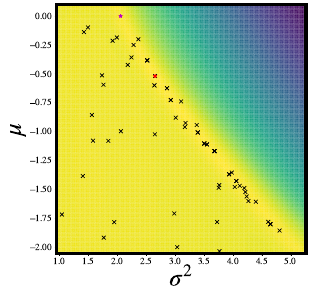}
\subcaption{$L=32$}
\end{minipage}\hfill
\begin{minipage}[t]{.15\linewidth}
\includegraphics[width=\linewidth]{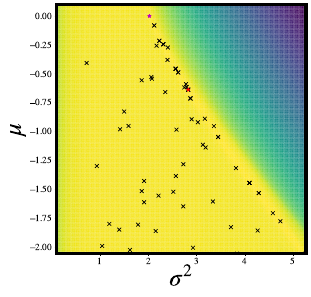}
\subcaption{$L=64$}
\end{minipage}\hfill
\begin{minipage}[t]{.15\linewidth}
\includegraphics[width=\linewidth]{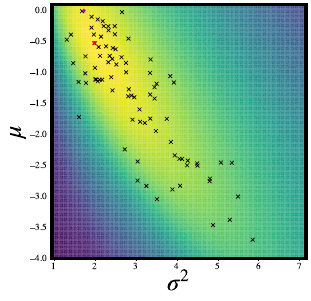}
\subcaption{$L=2$}
\end{minipage}\hfill
\begin{minipage}[t]{.15\linewidth}
\includegraphics[width=\linewidth]{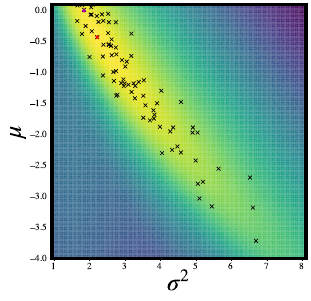}
\subcaption{$L=4$}
\end{minipage}\hfill
\begin{minipage}[t]{.15\linewidth}
\includegraphics[width=\linewidth]{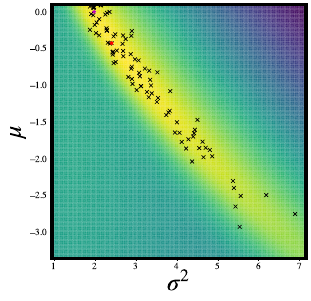}
\subcaption{$L=8$}
\end{minipage}\hfill
\begin{minipage}[t]{.15\linewidth}
\includegraphics[width=\linewidth]{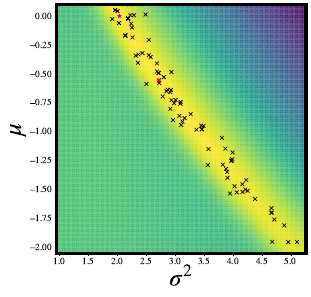}
\subcaption{$L=16$}
\end{minipage}\hfill
\begin{minipage}[t]{.15\linewidth}
\includegraphics[width=\linewidth]{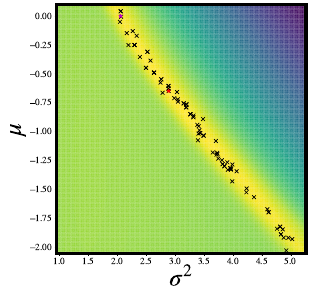}
\subcaption{$L=32$}
\end{minipage}\hfill
\begin{minipage}[t]{.15\linewidth}
\includegraphics[width=\linewidth]{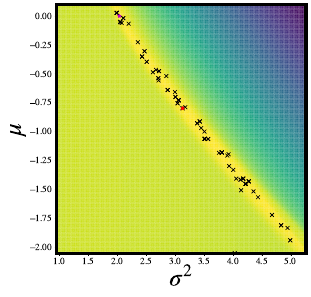}
\subcaption{$L=64$}
\end{minipage}\hfill
\begin{minipage}[t]{.15\linewidth}
\includegraphics[width=\linewidth]{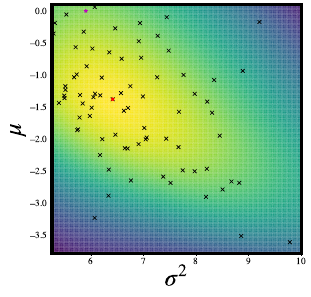}
\subcaption{$L=2$}
\end{minipage}\hfill
\begin{minipage}[t]{.15\linewidth}
\includegraphics[width=\linewidth]{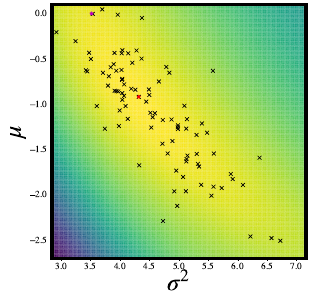}
\subcaption{$L=4$}
\end{minipage}\hfill
\begin{minipage}[t]{.15\linewidth}
\includegraphics[width=\linewidth]{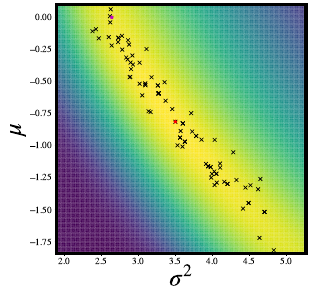}
\subcaption{$L=8$}
\end{minipage}\hfill
\begin{minipage}[t]{.15\linewidth}
\includegraphics[width=\linewidth]{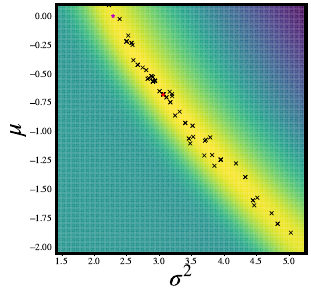}
\subcaption{$L=16$}
\end{minipage}\hfill
\begin{minipage}[t]{.15\linewidth}
\includegraphics[width=\linewidth]{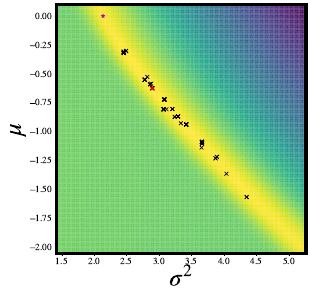}
\subcaption{$L=32$}
\end{minipage}\hfill
\begin{minipage}[t]{.15\linewidth}
\includegraphics[width=\linewidth]{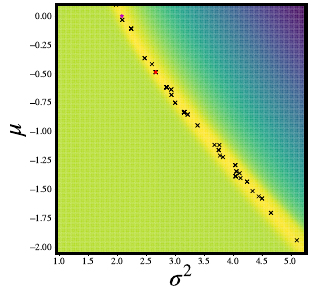}
\subcaption{$L=64$}
\end{minipage}\hfill
\caption{Log hyper-posterior. The red points indicate location of maxima, and the magenta points indicate the location of the maxima constrained along $\mu=0$ (up to a discrete grid of $200 \times 200$). The black points indicate samples obtained through MH. (a - f) \texttt{Sine}. (g - f) \texttt{Smooth XOR}. (m - r) \texttt{Snelson}.}
\label{fig:lpost}
\end{figure*}
\begin{figure*}[!htb]
\centering
\begin{minipage}[t]{.15\linewidth}
\includegraphics[width=\linewidth]{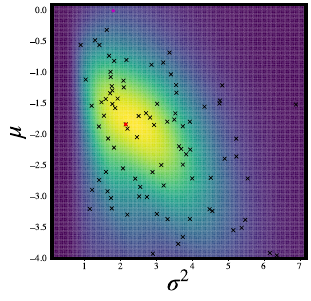}
\subcaption{$L=2$}
\end{minipage}\hfill
\begin{minipage}[t]{.15\linewidth}
\includegraphics[width=\linewidth]{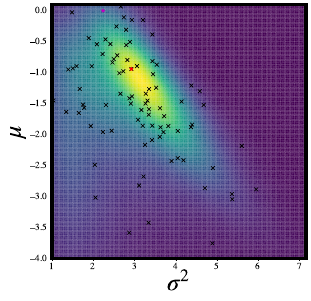}
\subcaption{$L=4$}
\end{minipage}\hfill
\begin{minipage}[t]{.15\linewidth}
\includegraphics[width=\linewidth]{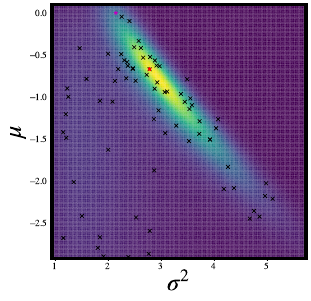}
\subcaption{$L=8$}
\end{minipage}\hfill
\begin{minipage}[t]{.15\linewidth}
\includegraphics[width=\linewidth]{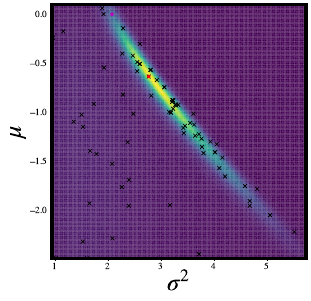}
\subcaption{$L=16$}
\end{minipage}\hfill
\begin{minipage}[t]{.15\linewidth}
\includegraphics[width=\linewidth]{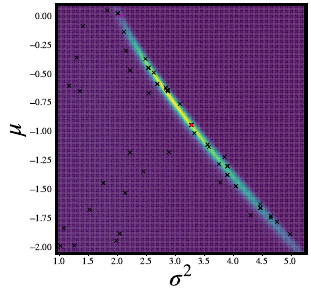}
\subcaption{$L=32$}
\end{minipage}\hfill
\begin{minipage}[t]{.15\linewidth}
\includegraphics[width=\linewidth]{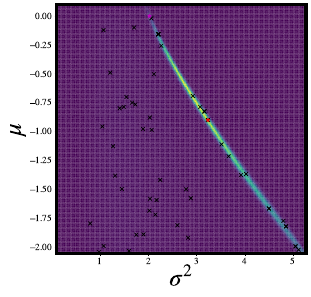}
\subcaption{$L=64$}
\end{minipage}\hfill
\begin{minipage}[t]{.15\linewidth}
\includegraphics[width=\linewidth]{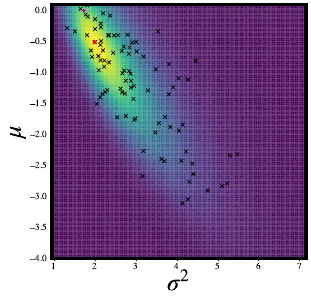}
\subcaption{$L=2$}
\end{minipage}\hfill
\begin{minipage}[t]{.15\linewidth}
\includegraphics[width=\linewidth]{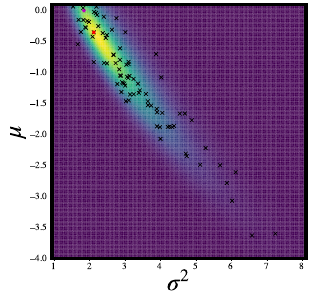}
\subcaption{$L=4$}
\end{minipage}\hfill
\begin{minipage}[t]{.15\linewidth}
\includegraphics[width=\linewidth]{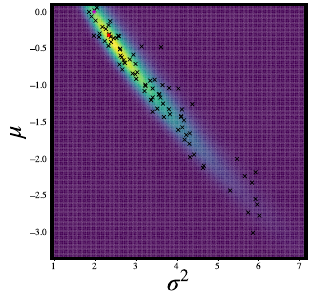}
\subcaption{$L=8$}
\end{minipage}\hfill
\begin{minipage}[t]{.15\linewidth}
\includegraphics[width=\linewidth]{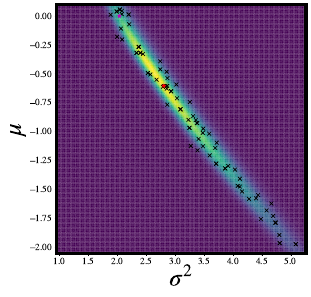}
\subcaption{$L=16$}
\end{minipage}\hfill
\begin{minipage}[t]{.15\linewidth}
\includegraphics[width=\linewidth]{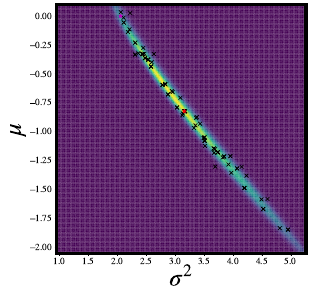}
\subcaption{$L=32$}
\end{minipage}\hfill
\begin{minipage}[t]{.15\linewidth}
\includegraphics[width=\linewidth]{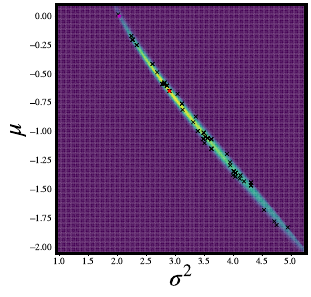}
\subcaption{$L=64$}
\end{minipage}\hfill
\begin{minipage}[t]{.15\linewidth}
\includegraphics[width=\linewidth]{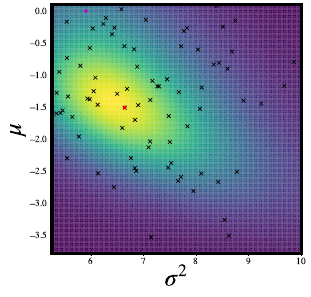}
\subcaption{$L=2$}
\end{minipage}\hfill
\begin{minipage}[t]{.15\linewidth}
\includegraphics[width=\linewidth]{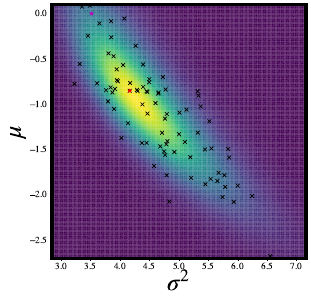}
\subcaption{$L=4$}
\end{minipage}\hfill
\begin{minipage}[t]{.15\linewidth}
\includegraphics[width=\linewidth]{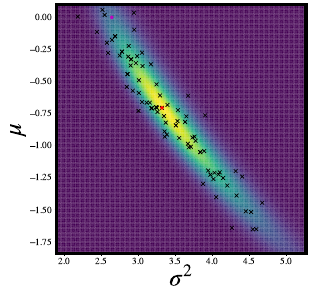}
\subcaption{$L=8$}
\end{minipage}\hfill
\begin{minipage}[t]{.15\linewidth}
\includegraphics[width=\linewidth]{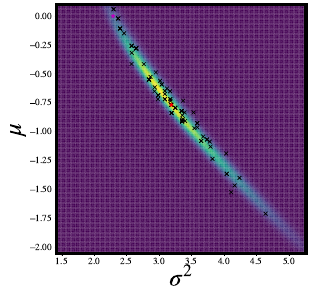}
\subcaption{$L=16$}
\end{minipage}\hfill
\begin{minipage}[t]{.15\linewidth}
\includegraphics[width=\linewidth]{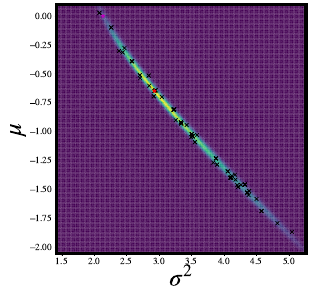}
\subcaption{$L=32$}
\end{minipage}\hfill
\begin{minipage}[t]{.15\linewidth}
\includegraphics[width=\linewidth]{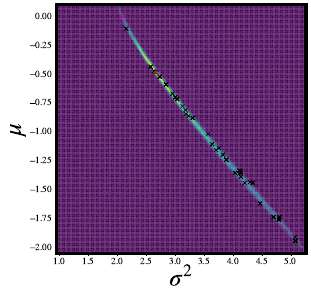}
\subcaption{$L=64$}
\end{minipage}\hfill
\caption{Hyper-posterior. The red points indicate location of maxima, and the magenta points indicate the location of the maxima constrained along $\mu=0$ (up to a discrete grid of $200 \times 200$). The nlack point indicates samples obtained through MH. (a - f) \texttt{Sine}. (g - f) \texttt{Smooth XOR}. (m - r) \texttt{Snelson}.}
\label{fig:post}
\end{figure*}

At the risk of overfitting and discarding uncertainties, the pragmatic Bayesian may be tempted to replace the marginalised model in~\eqref{eq:predictive} with the GP evaluated at the MLE or MAP. Such approaches are argued to be sensible approximations when the hyper-posterior is well-concentrated. Together with the likelihood, the hyper-posterior indeed becomes highly concentrated as the model depth increases, as shown in Figure~\ref{fig:lpost}. It is therefore of interest to compare the performance of MLE, MAP and marginalised models.

We use Metropolis-Hastings (MH) to approximately sample from the hyper-posterior, and for the MAP estimate we take the sample obtained through MH with the largest density under the hyper-posterior. To give a sense of the hyper-posterior, we also provide plots of the hyper-posterior and samples obtained with MH in Figures~\ref{fig:lpost} and~\ref{fig:post}. For the prior, we use a Gaussian distribution for $\mu$ and an inverse-gamma distribution for $\sigma^2$ with $\mu$ independent of $\sigma^2$. Full details of the MH settings and prior parameters are given in Appendix~\ref{app:mcmc}.

\subsection{Qualitative observations}
\label{sec:qualitative}
\subsubsection{Kernel expressivity with respect to hyperparameters}
It is interesting to note that the orange (MAP, $\mu=0$) and blue (MLE, $\mu=0$) curves in Figure~\ref{fig:performance} are very closely aligned. This is due to the simplicity of the way the single hyperaparameter $\sigma^2$ interacts with the kernel. As in~\eqref{eq:arc-cosine}, $\sigma^2$ only scales the kernel. In the case of noiseless GP regression (i.e. $s=0$ in~\eqref{eq:gp_pred}) with $\mu=0$, \emph{the mean of the posterior predictive is not affected by $\sigma^2$}. 

More generally when $s$ is small, we do not expect the mean to change much with $\sigma^2$ by a simple perturbation analysis. Let $c$ be as defined in~\eqref{eq:arc-cosine}, involving the products of the weight variances in each layer. Let $\ranvec{\overline{f}}_{*,c_1}$ and $\ranvec{\overline{f}}_{*,c_2}$ be the means of the posterior predictive distributions with $c=c_1$ and $c=c_2$. Let $K_{X,X} = K(\ranmat{X}^{(0)},\ranmat{X}^{(0)})/c^2$ and $K_{*,X} = K(\ranmat{X}_*, \ranmat{X}^{(0)})/c^2$ (neither of which depend on $s$, $c$ or $\sigma$). Then for all vector and induced matrix $p$-norms $\Vert \cdot \Vert$ and $p$-condition numbers $\kappa(\cdot)$,

\allowdisplaybreaks
\begin{align*}
\Vert \ranvec{\overline{f}}_{*,c_1} - \ranvec{\overline{f}}_{*,c_2} \Vert &= \Big\Vert K_{*,X} \Big( [K_{X,X} + \frac{s^2}{c_1^2} I]^{-1} \ranvec{y} - [K_{X,X} + \frac{s^2}{c_2^2} I]^{-1} \ranvec{y} \Big) \Big \Vert \\
&\leq \Big\Vert  K_{*,X} \Big\Vert  \Big \Vert \Big( [K_{X,X} + \frac{s^2}{c_1^2} I]^{-1}  \ranvec{y} - K_{X,X}^{-1}  \ranvec{y} + K_{X,X}^{-1}  \ranvec{y} - [K_{X,X} + \frac{s^2}{c_2^2} I]^{-1} \ranvec{y} \Big) \Big \Vert  \\ 
&\leq 2 \Big\Vert  K_{*,X} \Big\Vert  \Big \Vert  K_{X,X}^{-1} \ranvec{y} \Big\Vert \max_{c \in \{c_1, c_2\}} \Bigg\{ \frac{\kappa ( K_{X,X} )}{1 - \kappa ( K_{X,X} ) \frac{s^2 \Vert I \Vert}{c^2 \Vert K_{X,X} \Vert }} \frac{s^2 \Vert I \Vert}{c^2 \Vert K_{X,X} \Vert } \Bigg\} \\
&= 2 \Big\Vert  K_{*,X} \Big\Vert  \Big \Vert  K_{X,X}^{-1} \ranvec{y} \Big\Vert \max_{c \in \{c_1, c_2\}} \Bigg\{ \frac{  \Vert K_{X,X}^{-1} \Vert}{1 - \Vert K_{X,X}^{-1} \Vert \frac{s^2 }{c^2 }} \frac{s^2 }{c^2 } \Bigg\} \\
& \leq 2 s^2 \Big\Vert  K_{*,X} \Big\Vert  \Big \Vert \ranvec{y} \Big\Vert \max_{c \in \{c_1, c_2\}} \Bigg\{ \frac{ \lambda^2 }{c^2  - s^2 \lambda } \Bigg\},
\end{align*}
where $\lambda$ is the smallest eigenvalue of $K_{X,X}$, provided that $\frac{s^2}{c^2} \Vert K_{X,X}^{-1} \Vert  < 1 $. Even if $s$ is not too small compared to $c$, for \emph{random} $\ranmat{X}^{(0)}$, we expect $\lambda$ to decrease in the number of rows by a non-rigourous application of results from random matrix theory.

\subsubsection{Textbook overfitting}
A somewhat suprising observation is that the MSE curves in Figure~\ref{fig:performance} are consistent with overfitting. Using depth as a proxy for model complexity, models with $\mu \neq 0$ achieve a lower test error up until a certain level depth, at which point they start to overfit to the training data. This is noteworthy because all of the MLE and MAP models are simply GPs with $2$ hyperparameters. If we are to understand depth as a measure of complexity in this context, complexity only increases through the compositions of kernel functions. Furthermore, in \texttt{Smooth XOR} and \texttt{Snelson}, overfitting appears to be mitigated through the regularisation effect of constraining $\mu = 0$. In \texttt{Sine}, which is a very simple regression problem and therefore amenable to severe regularisation, models with $\mu=0$ outperform or are competitive with models with $\mu \neq 0$ for all depths.

\subsubsection{Ill-conditioning}
\label{sec:illcond}
The marginal likelihoods in Figure~\ref{fig:ll} appear to become increasingly concentrated along a diagonal as $L$ increases. This suggests that deeper models are more difficult to optimise than shallow models, a fact that is shared with finite-width models trained using gradient-based optimisers. 

The ill-conditioning made it difficult for our simple MH sampler to propose sensible jumps in deep models. This likely contributed to the performance curve of the marginalised model in Figure~\ref{fig:performance} increasing with depth. Future work involving hyperparameter marginalisation should look into a more advanced integration method.

\section{Conclusion}
\subsection{Summary of contributions}
While it has long been known that random MLPs with iid parameters converge to GPs, it has previously been assumed that the means of the parameters are zero. By generalising the parameter prior to allow for non-zero means, we are able to add extra flexibility to the limiting model. We evaluated the limiting kernel of deep networks with such priors.

We obtain a level II inference scheme by replacing the iid prior with a more general RCE prior. The RCE prior arises naturally due to the permutation symmetry in the MLP, and also has connections to networks trained using gradient-based optimisers.

When the means and variances are constrained to be equal in every layer, our results allow for the visualisation of the marginal likelihood and hyper-posterior in a $2$D plane. In our experiments, the additional mean parameter allowed for a more expressive kernel, but was also susceptible to overfitting. We also observed that the marginal likelihood becomes more ill-conditioned as the depth of the network increases.

\subsection{Discussion and future work}
While zero-mean limiting kernels have a fixed point in depth after accounting for scaling, Figures~\ref{fig:iid_kernel}(b, c) suggest that this fixed point may be avoided or non-trivial in models with non-zero means. We would like to explore this idea analytically in future. 

It is easy to make models with unconstrained means overfit as shown in Figure~\ref{fig:performance}, regardless of whether they are obtained through maximum-likelihood or are Bayesian and contain a degree of regularisation. However, when the means are constrained to be zero, it appears more difficult to overfit. Perhaps this insight can be used to better understand the mysterious generalisation properties of finite width networks trained using gradient-based optimisers.

Our RCE prior allows for dependencies of parameters inside layers, but we still assumed independence between layers~\eqref{eq:factor}. It remains open as to how one might introduce dependencies between layers while maintaining a degree of analytical tractability. It may be useful to apply the notion of partial exchangeability to other layer types. For example, it would be natural to study convolutional networks with exchangeable filters in each layer.

We evaluated the kernel when $\mu \neq 0$ and the activation functions are LReLU. Kernels for other activation functions such as the softplus~\citep{nair2010rectified}, ELU~\citep{clevert2015fast} or Swish~\citep{ramachandran2017swish} may be evaluated in the future. A comparison of the resultant kernels may help in understanding the biases of finite-width networks.

\subsubsection*{Acknowledgements}
Russell Tsuchida and Fred Roosta gratefully acknowledge the generous support given by the Australian Research Council Centre of Excellence for Mathematical \& Statistical Frontiers(ACEMS). Fred Roosta was partially supported by the Australian Research Council through a Discovery Early Career Researcher Award (DE180100923). This material is based on research partially sponsored by DARPA and the Air Force Research Laboratory under agreement number FA8750-17-2-0122.

\newpage
\onecolumn
\appendix
\section{Metropolis-Hastings settings and hyperprior}
\label{app:mcmc}
\subsection*{Hyper-prior}
We used the hyper-prior
\begin{equation*}
\mu \sim \mathcal{N}\big( -1, 2 \big), \qquad \sigma^2 \sim \text{Inv-Gamma}\big( 2.5, 6\big),
\end{equation*}
with independence between $\mu$ and $\sigma^2$.

\subsection*{Metropolis-Hastings settings}
We use a no-frills implementation of MH with a fixed bivariate Gaussian proposal distribution. The mean of the proposal distribution is given by the current point. The covariance matrix of the proposal distribution contains diagonal elements $\Sigma_{11} = 2.38$, $\Sigma_{22} = 4.76$ and off-diagonal element $\Sigma_{12} = -0.9 \sqrt{\Sigma_{11} \Sigma_{12}}$, roughly following the diagonal shape of the target distribution. Here $\Sigma_{11}$ refers to the variance of $\mu$ and $\Sigma_{22}$ refers the the variance of $\sigma^2$. We initialised the MH sampler at the MAP  over a $200 \times 200$ grid as shown in Figure~\ref{fig:lpost}, and ran the sampler for a burn-in period of $20$ samples. We kept every $20$th sample, and collected a total of $100$ samples. 

\vskip 0.2in
\bibliography{main}

\end{document}